\documentclass{investor-report}
\PassOptionsToPackage{numbers,sort&compress}{natbib}
\usepackage[T1]{fontenc}
\usepackage[utf8]{inputenc}
\usepackage{amsmath,amssymb,booktabs,graphicx,tabularx,array}
\usepackage{microtype,xcolor,hyperref,url,capt-of}
\hypersetup{colorlinks=true,citecolor=WeChatDeepGreen,linkcolor=WeChatDeepGreen,urlcolor=WeChatGreen,
 pdftitle={Investor's Last Exam: Can AI Agents Do Equity Investment Without Seeing the Future?},pdfauthor={Huangfei Jiang, Anning Hu, Kai Sheng, Yuchen Cai,
Rui Ge, Xiangyu Kong, Jianghao Lin, Shuzi Chen, Jiyi Weng,
Junchi Yan}}
\newcommand{\ILE}{\textsc{ILE}}
\newcommand{\code}[1]{\texttt{#1}}
\graphicspath{{figures/}}
\newcolumntype{Y}{>{\raggedright\arraybackslash}X}
\title{Investor's Last Exam:\\Can AI Agents Do Equity Investment\\Without Seeing the Future?}
\author{%
\begin{minipage}{0.98\textwidth}
\centering
\normalfont\normalsize
Huangfei Jiang$^{1*}$,
Anning Hu$^{1,2*}$,
Kai Sheng$^{1,3*}$,
Yuchen Cai$^{1,4}$,
Rui Ge$^{1}$,\\[2pt]
Xiangyu Kong$^{5}$,
Jianghao Lin$^{3}$,
Shuzi Chen$^{5}$,
Jiyi Weng$^{6}$,
Junchi Yan$^{2,\dagger}$
\par\vspace{6pt}

{\small
$^{1}$Prophetlab \quad
$^{2}$RethinkLab \quad
$^{4}$USTC \quad \\[2pt]
$^{3}$Antai College of Economics and Management, SJTU\\[2pt]
$^{5}$Foresight Fund \quad
$^{6}$IvyRock Asset Management
\par}

\vspace{5pt}
{\footnotesize
$^{*}$Equal contribution \quad
$^{\dagger}$Corresponding author
\par}
\end{minipage}%
}

\usepackage{needspace,enumitem}
\abstract{\small Financial agents must form forecasts and investment judgments from decision-time evidence, with predictions evaluated against subsequent outcomes. Existing benchmarks assess financial knowledge, research workflows, or forecasting, but evaluating the full research process requires a strictly frozen information set and professional evaluation criteria. We introduce \textbf{Investor's Last Exam (\ILE)}, an equity-research benchmark led by experienced buy-side investors and fund managers. Its three components are a versioned financial database spanning 3.43 million document versions and nine source categories, supplemented by structured market data, with validated timestamps governing retrieval and computation; a task taxonomy organized around professional analyst deliverables; and an expert-led framework linking scoring criteria to sources, calculations, and assumptions through financial dependency graphs. Evaluation spans 86 assignments and jointly assesses research quality, forecast accuracy, and simulated investment performance. GPT-6 Astra and Claude Opus 5.5 lead overall at 68.1 and 66.9 out of 100, yet even Astra scores only 52.5 on Forecast \& Model. Our results show that even state-of-the-art models fail to translate overall research capability into predictive performance. \ILE shifts the evaluation unit from isolated capabilities to the evidence--assumption--judgment chain, allowing researchers to locate where analytical advantages fail to translate into forecasts and investment judgments, and providing a traceable, diagnostic foundation for training and evaluating financial agents.}
\DeclareRobustCommand{\frontlinkicon}[1]{%
  \raisebox{-0.3em}{\includegraphics[width=0.025\linewidth]{#1}}}
\renewcommand{\checkdata}[2][]{\addtolist[#1]{#2}{\checkdatalist}{\checkdataformat}{\par}}
\checkdata[\frontlinkicon{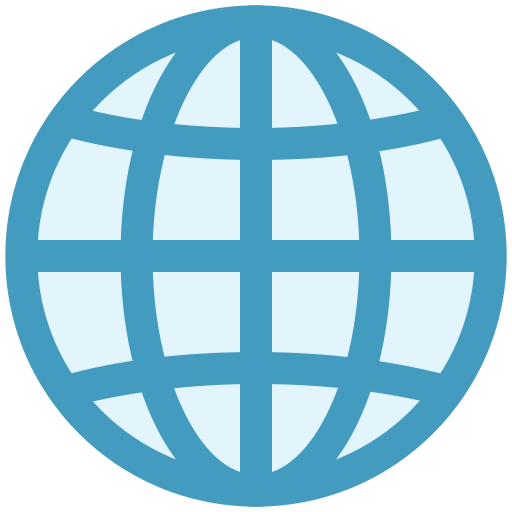}~~Project Page]{\url{https://prophet-lab.cn/benchmarks/investor-last-exam-v1}}
\checkdata[\frontlinkicon{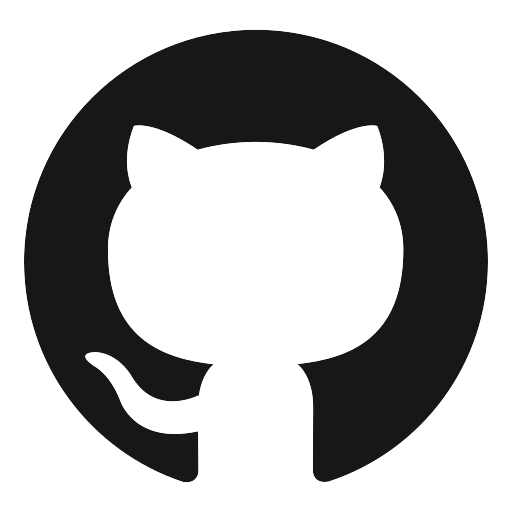}~~Code]{\url{https://github.com/Prophetlab/Investor-Last-Exam}}
\checkdata[\frontlinkicon{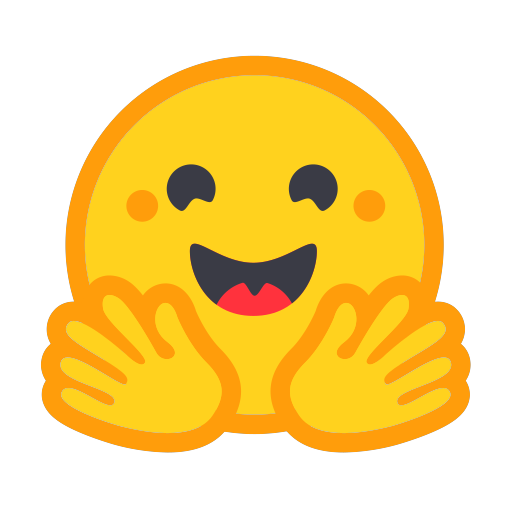}~~Dataset]{\url{https://huggingface.co/datasets/HenryJ12/InvestorLastExam}}

\setlist{topsep=3pt,itemsep=2pt,parsep=0pt,leftmargin=1.8em}
\let\reportsection\section
\renewcommand{\section}{\Needspace*{5\baselineskip}\reportsection}

\begin{document}
\raggedbottom
\enlargethispage{10mm}
\maketitle
\suppressfloats[t]

\begingroup
\setlength{\intextsep}{4pt}
\noindent\begin{minipage}{\linewidth}
\centering
\includegraphics[width=.80\linewidth]{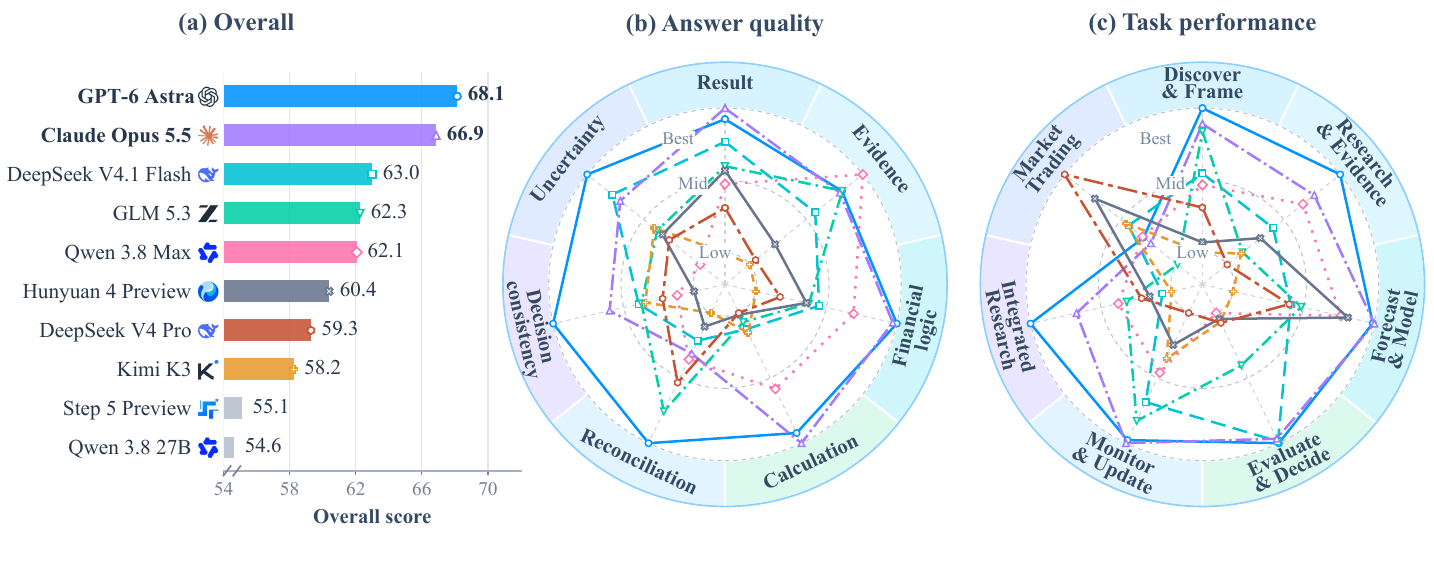}
\captionof{figure}{\textbf{Overall performance and relative strengths.} (a) Overall scores; (b) answer-quality dimensions; (c) professional task categories. Scores use GPT-6 Astra grading. Each radar axis spans the lowest (inner ring) to highest (outer ring) score among highlighted models. Table~\ref{tab:results} reports the detailed scores.}
\label{fig:diagnostics}
\end{minipage}
\endgroup

\clearpage

\begin{figure}[!ht]
\centering
\includegraphics[width=\linewidth]{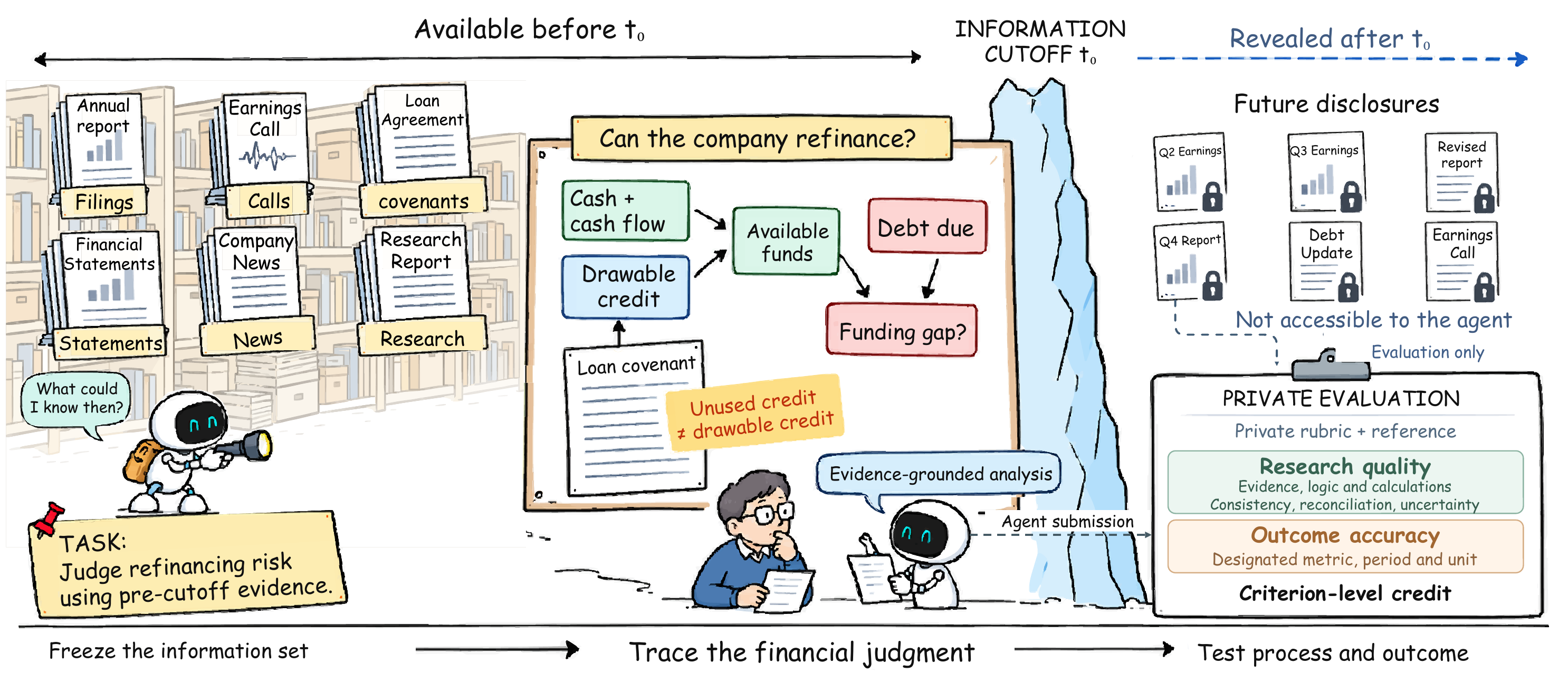}
\caption{\textbf{Cutoff-constrained financial research and private evaluation.} The agent analyzes sources available before $t_0$. In the refinancing example, drawable credit depends on loan covenants. Private references and rubrics assess research quality and designated outcomes; later disclosures are accessible only to evaluation.}
\label{fig:overview}
\end{figure}

\section{Introduction}\label{sec:intro}
Buy-side equity research informs capital allocation decisions~\citep{cfa2026equityvaluation}. Researchers analyze a company's operations, competitive position, and financial condition, model its future performance, and assess what these expectations imply for value, risk, and investment decisions. This process, from understanding a business to forming an investment thesis, is central to professional equity analysis. However, question answering, retrieval, or isolated computational skills are insufficient to characterize AI's competence in this work~\citep{2023_FinanceBench,2025_FinanceAgentBenchmark,2025_FinSearchComp}. Professional equity research is difficult to evaluate precisely because the quality of its judgments is not determined by any single capability, but by whether the analytical chain is coherent, the evidence is sufficient, and the forecasts withstand subsequent verification, attributes that are embedded in the real research process rather than capturable by isolated tasks.

Existing financial benchmarks mostly cover isolated stages of the research workflow. FinanceBench~\citep{2023_FinanceBench} focuses on document QA, Finance Agent~\citep{2025_FinanceAgentBenchmark} and FinSearchComp~\citep{2025_FinSearchComp} introduce agentic search, and FinBench~\citep{ghosh2026finbench} begins to evaluate forecasting and outcome validation. FinanceGym~\citep{xiao2026financeharness} further combines thesis-driven reports, expert-validated rubrics, and a publication-date-filtered web corpus. However, publication-date filtering remains a retrieval-level heuristic rather than a system-level constraint. It provides only an approximate temporal boundary, and page metadata can be misaligned with content updates, so post-cutoff information may still enter retrieval, computation, and scoring. Consequently, none of these benchmarks enforces version-level availability across the full research pipeline, leaving look-ahead leakage and evidence provenance uncontrolled. What remains missing is an evaluation unit that connects evidence, assumptions, and conclusions under a strict cutoff.

To this end, we introduce I\textbf{nvestor's Last Exam (\ILE)}, a benchmark for professional investment research. It evaluates whether AI agents can work within a realistic information boundary along the full analyst workflow, processing large-scale, interconnected financial data and documents to complete long-horizon tasks from evidence retrieval and assumption modeling to forecasting, valuation, and investment decisions. \ILE\ has three components. First, a versioned database of 3.43 million document versions across nine financial data categories (1.52\,TB) that supports retrieval, computation, and evidence tracing within the pre-cutoff information set, upgrading the cutoff from retrieval-level date filtering to a system-level constraint across data access, computation, and rubric review. Second, a professional task taxonomy of five categories and twenty subtypes plus integrated assignments, which decomposes the analyst workflow into scorable professional judgments such as accounting reconciliation, assumption-sensitive forecasting, and evidence-based investment decisions, rather than isolated capability tests. Third, an expert--AI pipeline that links scoring propositions to sources, calculations, and assumptions via financial dependency graphs, making reasoning paths for difficult tasks reviewable and traceable; experts provide questions and judgments, agents assemble retrievable and computable evidence, multiple models assist in review, and experts finalize the rubrics. Together, these components connect professional research to reviewable evidence and financial judgment, with expert criteria assessing both the analysis and designated outcomes to locate where agents succeed or fail. Table~\ref{tab:benchmark-comparison} summarizes how existing benchmarks differ along these dimensions.

\begin{table}[!t]
\centering\footnotesize
\setlength{\tabcolsep}{1.0pt}
\renewcommand{\arraystretch}{1.08}
\caption{\textbf{Design and evaluation features of financial benchmarks.} Year denotes the first public release.}
\label{tab:benchmark-comparison}
\begin{tabularx}{.96\linewidth}{@{}>{\raggedright\arraybackslash}p{.23\linewidth}>{\centering\arraybackslash}p{.055\linewidth}*{8}{>{\centering\arraybackslash}X}@{}}
\toprule
& & \multicolumn{3}{c}{Data sources} & \multicolumn{2}{c}{Agent tasks} & \multicolumn{3}{c}{Evaluation} \\
\cmidrule(lr){3-5}\cmidrule(lr){6-7}\cmidrule(l){8-10}
Benchmark & Year & \shortstack{Issuer\\docs} & \shortstack{News /\\web} & \shortstack{Market\\data} & \shortstack{Agent\\loop} & \shortstack{Forecast\\tasks} & \shortstack{Source\\trace} & \shortstack{Cutoff\\control} & \shortstack{Process\\rubric} \\
\midrule
FinQA~\citep{2021_FinQA} & 2021 & \textcolor[HTML]{228B50}{$\checkmark$} & \textcolor[HTML]{C43C39}{--} & \textcolor[HTML]{C43C39}{--} & \textcolor[HTML]{C43C39}{--} & \textcolor[HTML]{C43C39}{--} & \textcolor[HTML]{228B50}{$\checkmark$} & \textcolor[HTML]{C43C39}{--} & \textcolor[HTML]{C43C39}{--} \\
FinanceBench~\citep{2023_FinanceBench} & 2023 & \textcolor[HTML]{228B50}{$\checkmark$} & \textcolor[HTML]{C43C39}{--} & \textcolor[HTML]{C43C39}{--} & \textcolor[HTML]{C43C39}{--} & \textcolor[HTML]{C43C39}{--} & \textcolor[HTML]{228B50}{$\checkmark$} & \textcolor[HTML]{C43C39}{--} & \textcolor[HTML]{C43C39}{--} \\
Finance Agent~\citep{2025_FinanceAgentBenchmark} & 2025 & \textcolor[HTML]{228B50}{$\checkmark$} & \textcolor[HTML]{228B50}{$\checkmark$} & \textcolor[HTML]{C43C39}{--} & \textcolor[HTML]{228B50}{$\checkmark$} & \textcolor[HTML]{C43C39}{--} & \textcolor[HTML]{228B50}{$\checkmark$} & \textcolor[HTML]{C43C39}{--} & \textcolor[HTML]{228B50}{$\checkmark$} \\
FinSearchComp~\citep{2025_FinSearchComp} & 2025 & \textcolor[HTML]{228B50}{$\checkmark$} & \textcolor[HTML]{228B50}{$\checkmark$} & \textcolor[HTML]{228B50}{$\checkmark$} & \textcolor[HTML]{228B50}{$\checkmark$} & \textcolor[HTML]{C43C39}{--} & \textcolor[HTML]{C43C39}{--} & \textcolor[HTML]{C43C39}{--} & \textcolor[HTML]{C43C39}{--} \\
StockBench~\citep{2025_StockBench} & 2025 & \textcolor[HTML]{C43C39}{--} & \textcolor[HTML]{228B50}{$\checkmark$} & \textcolor[HTML]{228B50}{$\checkmark$} & \textcolor[HTML]{228B50}{$\checkmark$} & \textcolor[HTML]{C43C39}{--} & \textcolor[HTML]{C43C39}{--} & \textcolor[HTML]{228B50}{$\checkmark$} & \textcolor[HTML]{C43C39}{--} \\
FinDeepForecast~\citep{li2026findeepforecast} & 2026 & \textcolor[HTML]{228B50}{$\checkmark$} & \textcolor[HTML]{228B50}{$\checkmark$} & \textcolor[HTML]{228B50}{$\checkmark$} & \textcolor[HTML]{228B50}{$\checkmark$} & \textcolor[HTML]{228B50}{$\checkmark$} & $\circ$ & \textcolor[HTML]{228B50}{$\checkmark$} & \textcolor[HTML]{C43C39}{--} \\
BigFinanceBench~\citep{wang2026bigfinancebench} & 2026 & \textcolor[HTML]{228B50}{$\checkmark$} & \textcolor[HTML]{228B50}{$\checkmark$} & \textcolor[HTML]{228B50}{$\checkmark$} & \textcolor[HTML]{228B50}{$\checkmark$} & \textcolor[HTML]{C43C39}{--} & \textcolor[HTML]{C43C39}{--} & \textcolor[HTML]{C43C39}{--} & \textcolor[HTML]{228B50}{$\checkmark$} \\
FinBench~\citep{ghosh2026finbench} & 2026 & \textcolor[HTML]{C43C39}{--} & \textcolor[HTML]{228B50}{$\checkmark$} & \textcolor[HTML]{228B50}{$\checkmark$} & \textcolor[HTML]{C43C39}{--} & \textcolor[HTML]{228B50}{$\checkmark$} & \textcolor[HTML]{C43C39}{--} & \textcolor[HTML]{228B50}{$\checkmark$} & \textcolor[HTML]{C43C39}{--} \\
FinanceGym~\citep{xiao2026financeharness} & 2026 & \textcolor[HTML]{C43C39}{--} & \textcolor[HTML]{228B50}{$\checkmark$} & \textcolor[HTML]{C43C39}{--} & \textcolor[HTML]{228B50}{$\checkmark$} & \textcolor[HTML]{228B50}{$\checkmark$} & \textcolor[HTML]{228B50}{$\checkmark$} & \textcolor[HTML]{228B50}{$\checkmark$} & \textcolor[HTML]{228B50}{$\checkmark$} \\
FrontierFinance~\citep{zhang2026frontierfinance} & 2026 & \textcolor[HTML]{228B50}{$\checkmark$} & \textcolor[HTML]{228B50}{$\checkmark$} & \textcolor[HTML]{228B50}{$\checkmark$} & \textcolor[HTML]{228B50}{$\checkmark$} & \textcolor[HTML]{C43C39}{--} & $\circ$ & $\circ$ & \textcolor[HTML]{228B50}{$\checkmark$} \\
\midrule
\textbf{ILE} & \textbf{2026} & 
\scalebox{1.6}{\textcolor[HTML]{228B50}{\boldmath$\checkmark$}} & 
\scalebox{1.6}{\textcolor[HTML]{228B50}{\boldmath$\checkmark$}} & 
\scalebox{1.6}{\textcolor[HTML]{228B50}{\boldmath$\checkmark$}} & 
\scalebox{1.6}{\textcolor[HTML]{228B50}{\boldmath$\checkmark$}} & 
\scalebox{1.6}{\textcolor[HTML]{228B50}{\boldmath$\checkmark$}} & 
\scalebox{1.6}{\textcolor[HTML]{228B50}{\boldmath$\checkmark$}} & 
\scalebox{1.6}{\textcolor[HTML]{228B50}{\boldmath$\checkmark$}} & 
\scalebox{1.6}{\textcolor[HTML]{228B50}{\boldmath$\checkmark$}} \\
\bottomrule
\end{tabularx}
\vspace{3pt}
\parbox{.96\linewidth}{\scriptsize \textcolor[HTML]{228B50}{$\checkmark$}: supported; $\circ$: partial support; \textcolor[HTML]{C43C39}{--}: not reported.}
\end{table}
\FloatBarrier


We conduct experiments on eleven systems. GPT-6 Astra leads overall at 68.1, followed by Claude Opus 5.5 at 66.9, but Astra scores 78.9 on Research \& Evidence and 77.6 on Evaluate \& Decide while earning only 52.5 on Forecast \& Model, indicating that evidence organization and investment judgment are comparatively mature, whereas forecast construction remains a clear weakness. This weakness directly manifests as reordering under predictive criteria: DeepSeek V4.1 Flash leads realized forecast-outcome criteria, and Kimi K3 earns higher simulated returns than Opus despite a lower research score. This indicates that supporting analysis, forecast outcomes, and investment performance do not share a single ordering, and that an aggregate score alone cannot characterize research capability. We further trace score differences to specific financial dependencies and errors: systems reconcile cash-flow identities but struggle to separate operating improvement from temporary cash movements; they produce coherent driver bridges and recommendations but fail to ensure that assumptions, accounting states, and outputs form a testable financial model; submission control and error recovery are additional failure sources; and expert refinement shows that some errors reflect ambiguous scoring requirements rather than model capability alone. Overall, failures concentrate in forecast assumptions, accounting scope, financial interpretation, and decision consistency rather than in retrieval or arithmetic, indicating that the bottleneck for financial agents has shifted from information access to assumption calibration and model consistency.

\Needspace*{8\baselineskip}
Our main contributions are as follows:
\begin{itemize}\setlength\itemsep{2pt}
    \item \textbf{A time-frozen research environment.} Built on a versioned corpus, it enforces source-version availability across retrieval and computation, so agents can access only pre-decision evidence and look-ahead leakage is suppressed at the mechanism level.
    \item \textbf{An expert-constructed task suite and evaluation framework.} Tasks are authored by buy-side experts, and each scored requirement is anchored to evidence, financial dependencies, and verifiable criteria, traceable to a specific source document or calculation.
    \item \textbf{A multidimensional evaluation of forecasting and investment research.} Spanning forecast formation, valuation judgment, and investment decisions, it links judgments to supporting analyses and realized outcomes, assessing both outcome accuracy and reasoning consistency.
\end{itemize}

\begin{figure}[!t]
\centering
\includegraphics[width=\linewidth]{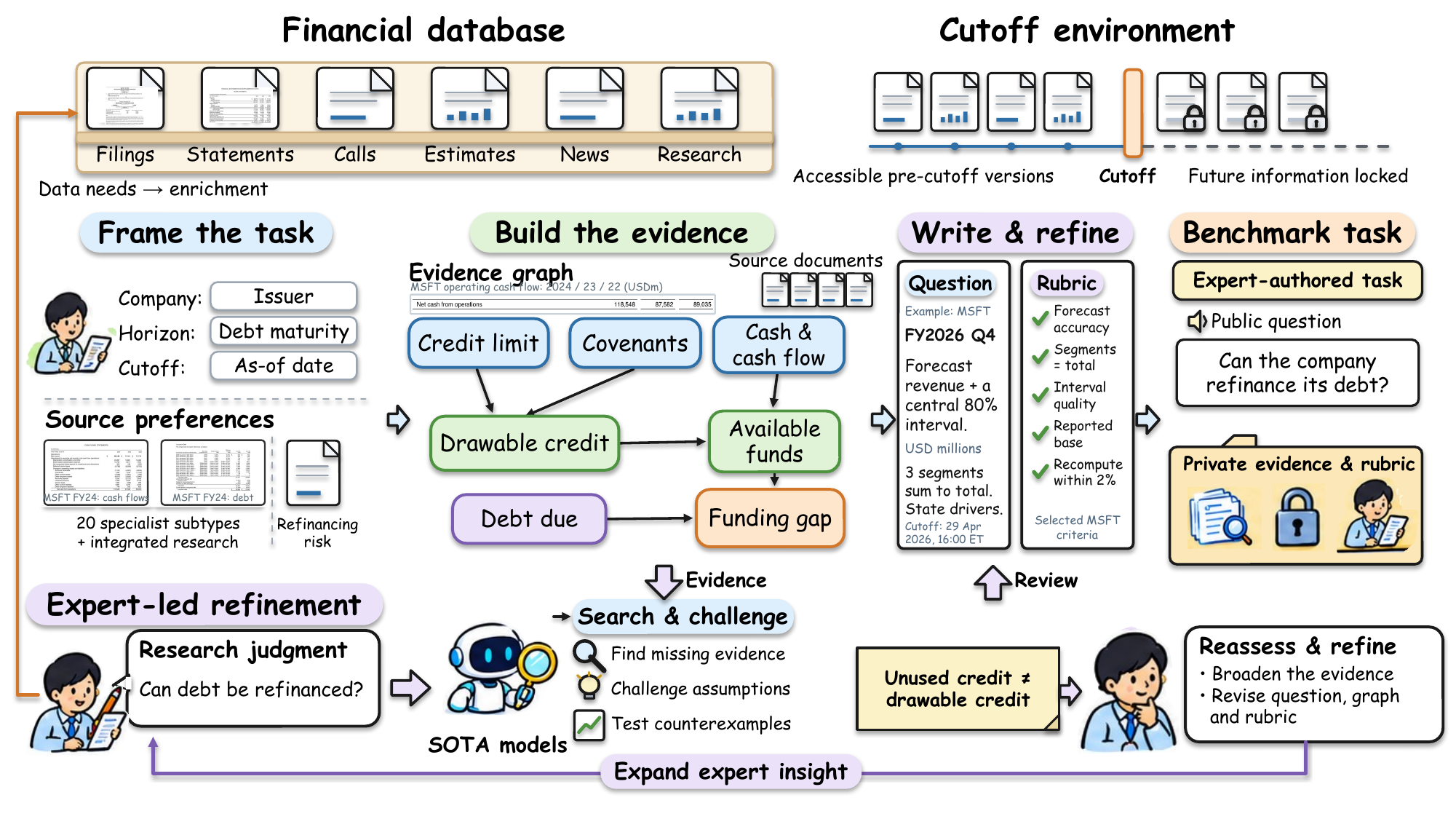}
\caption{\textbf{Expert-led construction with AI-assisted search and review.} Experts frame questions, build evidence-linked financial dependencies, and refine questions and rubrics. The refinancing graph illustrates a covenant constraint on drawable credit. Microsoft excerpts and the abbreviated O58 question and rubric show how reporting definitions enter task design. Reference analyses and rubrics remain private.}
\label{fig:construction}
\end{figure}
\section{Benchmark Design}\label{sec:environment}
The design goal of \ILE\ is to evaluate equity research under a strict information boundary: it uses professional research deliverables as task carriers and versioned evidence as the information boundary (Figure~\ref{fig:overview}), so that the evidence available to the agent at decision time and the financial requirements its answer must satisfy are constrained together. Section~\ref{versioned-evidence} defines versioned evidence and the information cutoff, Section~\ref{sec:tasks} presents the deliverable taxonomy and financial dependencies, and Section~\ref{sec:construction} details the expert-led construction framework that links scoring criteria to evidence, calculations, and assumptions.

\subsection{Financial Research Environment}
\label{versioned-evidence}

\begin{figure}[t]
\centering
\includegraphics[width=\linewidth]{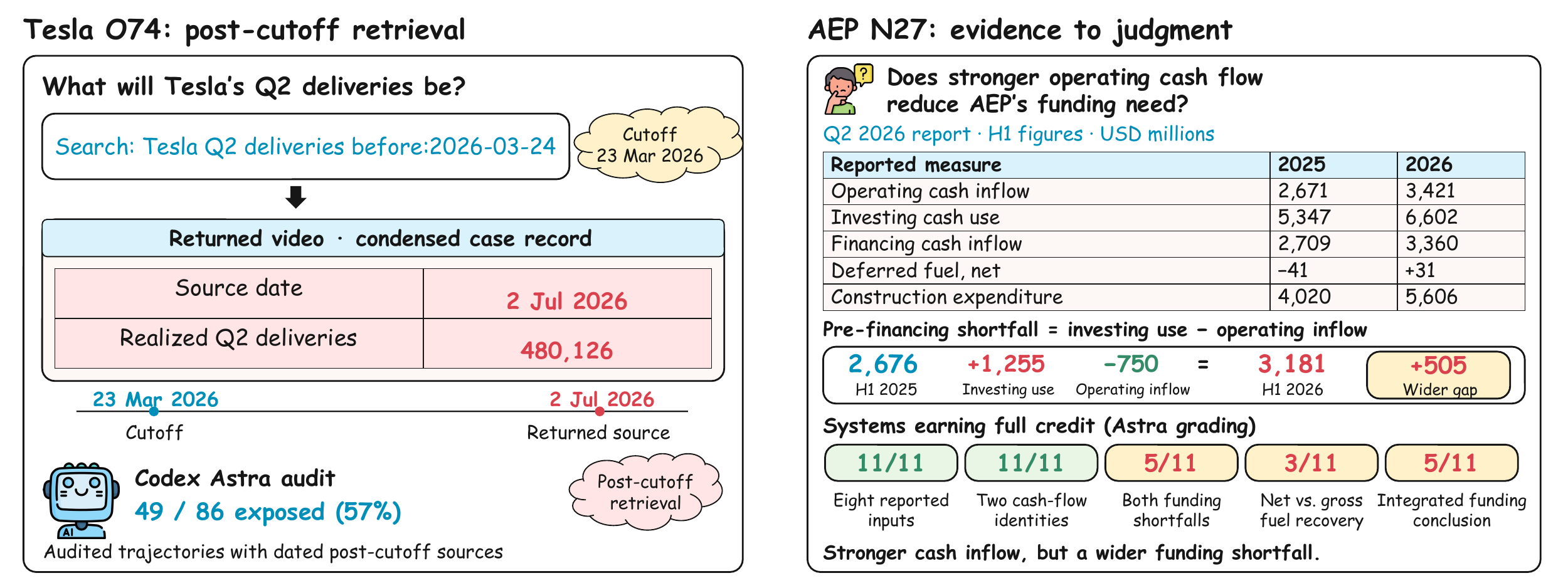}
\caption{\textbf{Information validity and financial interpretation.} Left: in Tesla task O74, GPT-6 Astra's public-web Codex search returns realized Q2 deliveries from a post-cutoff source despite a dated query. Right: higher AEP operating cash inflow coexists with a wider pre-financing shortfall. AEP amounts are USD millions; investing use and construction denote outflows. Each AEP badge counts how many of the 11 evaluated systems earn full credit on every criterion in the indicated group of this task's rubric, under GPT-6 Astra grading.}
\label{fig:research-cases}
\end{figure}

Merely constraining the query by date does not guarantee that the retrieved results stay within the specified cutoff. In the Tesla task O74 shown in Figure~\ref{fig:research-cases} (left), a GPT-6 Astra Codex trajectory issued a query containing \texttt{before:2026-03-24}, yet retrieved a source dated July~2 that reported realized Q2 deliveries. This is because a page's publication date need not reflect when its content actually became available, and missing metadata, updated pages, and retrospective reports can all bypass date filtering~\citep{2026_SearchDateLeakage_ACL}. Date filters constrain the query rather than the retrieved information, so a page can retain an early publication date while carrying later content.

To make the time constraint apply to the information itself, we build a continuously updated corpus that retains every historical version and assigns each version an availability time. As of 15 October 2026, the corpus contains 3,430,336 document versions and 1.52\,TB of material~\citep{prophetlab2026ile}, and is continually expanded with new disclosures and revisions while earlier versions are never overwritten. Sources fall into nine categories, covering financial reports, structured statements, earnings calls, investor materials, issuer events, analyst estimates, research reports, institutional holdings, and news. Structured market data support price, return, and volatility analysis, and coverage spans the US, mainland Chinese, Hong Kong, and other markets (Appendix~\ref{app:inventory}). Each version's availability time is established at collection time rather than inferred afterwards. Every immutable version records its publication or release time, timezone, and precision, together with supporting time evidence. Timestamp normalization uses a conservative availability boundary. For records with unverified timestamps, we first employ state-of-the-art models to infer and cross-validate their publication dates and identify potential subsequent updates. Records are excluded from task access only when a reliable availability time cannot be established, and each revision receives its own availability time.

Once established, availability time must take effect across the toolchain the agent actually uses. Each task defines a timezone-aware cutoff $t_c$ and authorized datasets; a version is eligible only if its availability time precedes $t_c$, regardless of its reporting period. The shared tool library enforces this boundary across the whole toolchain rather than at the search entry alone. Search, reads, downloads, and cached results share the same authorization and availability checks (Appendix~\ref{app:environment-contract}). \code{search} discovers eligible source versions and \code{read} returns their passages and tables. Structured-table tools query authorized records and freeze the results, so that later revisions cannot alter data already retrieved. \code{download} and \code{table\_download} retrieve eligible source files and export frozen results, respectively, both writing to the task workspace, where \code{bash} supports file inspection, numerical analysis, and model building with network access disabled.

Agents can therefore work autonomously with complete files and tables, leveraging subagent-assisted reading and context compression to build forecasts and reconciliations even when the required evidence exceeds the model's context window, while keeping all inputs within the cutoff. Source and version references link calculations to eligible evidence, tool records make the analysis inspectable, and \code{submit\_answer} validates the deliverable format. The evaluation side remains information-isolated, with reference analyses, rubrics, and later disclosures kept private from the agent. Forecast settlement then matches the target metric, period, and unit. Appendix~\ref{app:external-tools} evaluates the tool library on external financial tasks.

\subsection{Professional Task Taxonomy}\label{sec:tasks}

The taxonomy derives from the research practice of experienced buy-side analysts and fund managers. That practice is organized around deliverables, spanning four stages: screening opportunities, building evidence and forecasts, making investment recommendations, and monitoring theses. The taxonomy is structured around these same deliverables. Tasks are classified by their principal analyst deliverable (Table~\ref{tab:taxonomy}): five categories contain twenty subtypes, Integrated Research combines these functions into a full report, and Market Trading captures the ultimate downstream objective of investment research, targeting realized prices, returns, or volatility. A task has one category even when it involves several analytical operations, and Appendix~\ref{app:taxonomy} gives the detailed professional judgments and subtype definitions.

Financial difficulty in these tasks comes from the interdependence among measurements, assumptions, and decisions. Inputs must be comparable across reporting definitions; new evidence must justify the magnitude and scope of a forecast update; and an investment stance must follow from the model and identify the evidence that would reverse it. Plausible facts and local calculations alone do not satisfy these requirements, which also explains why the taxonomy does not classify tasks by topic. Two cases illustrate how these dependencies surface. Microsoft O58 shows why reporting scope matters, since cloud revenue spans the three segments and cannot be added as a fourth. AEP N27 tests whether higher operating cash flow implies lower financing needs once investment spending and cash-flow adjustments are considered. Section~\ref{sec:case-aep-main} traces the latter from question to evidence and criterion-level results, and Appendix~\ref{app:financial-cases} gives additional cases.

\begin{table}[!b]
\centering\small
\setlength{\tabcolsep}{4pt}
\renewcommand{\arraystretch}{1.08}
\caption{\textbf{Tasks organized by analyst deliverable.} Categories specify the work assigned; scoring dimensions assess requirements within its answer.}
\label{tab:taxonomy}
\begin{tabularx}{\linewidth}{@{}>{\raggedright\arraybackslash}p{.26\linewidth}Yr@{}}
\toprule
Task category & Representative deliverable & Tasks \\
\midrule
Discover \& Frame & Opportunity screen, catalyst map, or testable thesis & 10 \\
Research \& Evidence & Comparable historical series or reconciled facts & 13 \\
Forecast \& Model & Operating forecast, earnings bridge, or peer read-through & 13 \\
Evaluate \& Decide & Valuation, implied expectations, or investment action & 14 \\
Monitor \& Update & Event assessment, revised forecast, or thesis update & 17 \\
Integrated Research & Report linking forecasts, stance, risks, and thesis conditions & 11 \\
Market Trading & Price forecast or return/volatility ranking & 8 \\
\bottomrule
\end{tabularx}
\end{table}



\subsection{Expert-Led Task Construction}\label{sec:construction}

Building on the taxonomy, this section explains how tasks are constructed. Construction begins with experts fixing the task elements: drawing on their research practice, they specify the entity or universe, period, deliverable, and investment judgment (Figure~\ref{fig:construction}), where the deliverable and investment judgment directly correspond to the classification basis of the previous section. These questions in turn drive data development; when required source classes are missing, acquisition, parsing, and version tracking are expanded, yielding reusable coverage under the same information boundary.

The core of construction is to transform each task into a verifiable financial dependency graph that connects source-grounded facts, financial assumptions, intermediate calculations, and investment judgments through explicit financial relationships. While building the reference analysis, experts use \code{search} and \code{read} to select source passages and attach them to claims and calculations; each fact retains its issuer, period, unit, source version, and location, and relations express the financial rules connecting inputs and outputs, such as an accounting identity, revenue bridge, or peer read-through. This is precisely how the interdependence among measurements, assumptions, and decisions described in the previous section materializes within an individual task. AI-assisted investigation then searches for missing evidence, tests assumptions, and proposes counterexamples, with multiple state-of-the-art models reviewing drafts, but whether a suggestion is adopted is decided by the expert. The refinancing example illustrates this division of labor. Cash plus cash flow plus unused credit appears sufficient until a borrowing covenant is examined, at which point the expert adds the decisive dependency, namely whether credit is drawable in that state, to the question, reference analysis, and rubric.

Once construction is complete, each scoring criterion is linked to supporting evidence, a reproducible calculation, or an explicit financial judgment, making the evaluation traceable and verifiable. Experts then review cutoff validity, units, accounting scope, assumptions, and question--rubric alignment, while the rubric specifies weights, evidence, and tolerances or partial credit. Trial answers expose ambiguity and incomplete financial arguments, and authors trace discrepancies to the affected source, assumption, question, or scoring rule and revise that component. Expert review ultimately determines whether the scoring rules test the intended financial judgment, including the validity of accounting definitions, forecast assumptions, and reward incentives. AWS O15 offers an instance of calibration. Qwen 27B's initial revenue interval misses the realized outcome yet earns 15.8 points for being narrow; review specifies a central 80\% interval and replaces separate width credit with a joint accuracy--uncertainty score, and its revised interval also misses the outcome and earns zero interval credit (Appendix~\ref{app:case-aws}). The change makes uncertainty credit depend on both informativeness and predictive accuracy, and returns the scoring rules to the financial requirements set out in the previous section.

\begin{table}[!t]
\centering\footnotesize
\setlength{\tabcolsep}{2.5pt}
\renewcommand{\arraystretch}{1.12}
\caption{\textbf{Answer-based performance by scoring dimension (a) and task category (b).} Final answers are graded independently of submission acceptance, using GPT-6 Astra for semantic criteria. Panel (a) averages task-normalized dimension credit over applicable tasks; (b) averages complete task scores within each category. Overall weights tasks equally. Cost is answering-model API expenditure in USD per task; bold marks the highest displayed score.}
\label{tab:results}
\begin{tabular*}{\linewidth}{@{\extracolsep{\fill}}lrrrrrrrr@{}}
\toprule
\multicolumn{9}{@{}l}{\textbf{(a) Scoring dimensions: quality within answers}} \\
\addlinespace[2pt]
& & \multicolumn{7}{c}{Rubric dimensions (across applicable tasks)} \\
\cmidrule(l){3-9}
System & Overall & Result & Evidence & Logic & Calc. & Recon. & Decision & Uncert. \\
\midrule
GPT-6 Astra & \textbf{68.1} & 59.9 & 79.0 & \textbf{77.8} & 89.4 & \textbf{86.8} & \textbf{84.4} & \textbf{83.6} \\
Claude Opus 5.5 & 66.9 & \textbf{60.6} & 78.8 & 77.5 & \textbf{90.1} & 79.8 & 77.8 & 80.1 \\
DeepSeek V4.1 Flash & 63.0 & 58.5 & 76.2 & 72.0 & 82.5 & 78.8 & 74.5 & 81.0 \\
GLM 5.3 & 62.3 & 56.9 & 79.0 & 71.3 & 82.1 & 84.3 & 74.1 & 76.3 \\
Qwen 3.8 Max & 62.1 & 55.8 & \textbf{81.1} & 74.6 & 86.5 & 80.3 & 70.0 & 71.9 \\
Hunyuan 4 Preview & 60.4 & 56.7 & 72.1 & 71.1 & 81.6 & 77.7 & 68.0 & 75.8 \\
DeepSeek V4 Pro & 59.3 & 54.3 & 70.1 & 69.2 & 81.5 & 82.1 & 71.7 & 75.1 \\
Kimi K3 & 58.2 & 51.5 & 69.5 & 67.4 & 82.8 & 76.7 & 73.7 & 76.7 \\
Step 5 Preview & 55.1 & 51.8 & 71.9 & 65.1 & 79.3 & 68.2 & 61.8 & 75.5 \\
Qwen 3.8 27B & 54.6 & 47.3 & 68.6 & 63.4 & 78.5 & 74.9 & 69.1 & 69.1 \\
Ling 3.0 Flash & 42.6 & 43.9 & 47.9 & 45.3 & 60.3 & 53.3 & 54.1 & 40.9 \\
\bottomrule
\end{tabular*}
\par\vspace{7pt}
\begin{tabular*}{\linewidth}{@{\extracolsep{\fill}}lrrrrrrrr@{}}
\toprule
\multicolumn{9}{@{}l}{\textbf{(b) Task categories: performance across assignments}} \\
\addlinespace[2pt]
& \multicolumn{7}{c}{Professional task categories (full task scores)} & Resources \\
\cmidrule(lr){2-8}\cmidrule(l){9-9}
System & \shortstack{Discover\\\& Frame} & \shortstack{Research\\\& Evidence} & \shortstack{Forecast\\\& Model} & \shortstack{Evaluate\\\& Decide} & \shortstack{Monitor\\\& Update} & \shortstack{Integrated\\Research} & \shortstack{Market\\Trading} & \shortstack{API cost\\USD/task} \\
\midrule
GPT-6 Astra & \textbf{79.5} & \textbf{78.9} & 52.5 & \textbf{77.6} & 63.9 & \textbf{66.3} & 57.0 & 9.11 \\
Claude Opus 5.5 & 78.0 & 75.7 & \textbf{52.6} & 77.2 & \textbf{64.2} & 61.8 & 56.3 & 7.96 \\
DeepSeek V4.1 Flash & 73.4 & 70.8 & 45.8 & 77.4 & 60.5 & 53.5 & 58.5 & 0.36 \\
GLM 5.3 & 77.3 & 67.1 & 46.6 & 70.7 & 62.2 & 56.9 & 53.8 & 0.96 \\
Qwen 3.8 Max & 72.3 & 74.4 & 49.9 & 66.0 & 57.8 & 57.8 & 57.1 & 1.63 \\
Hunyuan 4 Preview & 66.9 & 69.3 & 50.5 & 66.5 & 55.3 & 54.7 & 61.7 & 0.29 \\
DeepSeek V4 Pro & 70.2 & 65.3 & 45.6 & 66.9 & 52.5 & 55.5 & \textbf{64.6} & 0.38 \\
Kimi K3 & 65.9 & 67.0 & 41.0 & 66.8 & 56.6 & 52.6 & 58.7 & 1.06 \\
Step 5 Preview & 69.0 & 64.1 & 43.9 & 54.9 & 58.1 & 49.0 & 44.0 & 0.72 \\
Qwen 3.8 27B & 60.7 & 60.3 & 44.5 & 62.1 & 53.9 & 46.7 & 53.7 & 0.32 \\
Ling 3.0 Flash & 46.5 & 41.8 & 38.5 & 49.6 & 39.7 & 34.5 & 50.9 & 0.04 \\
\bottomrule
\end{tabular*}
\end{table}

\Needspace*{8\baselineskip}
\section{Evaluation and Results}\label{sec:results}
\subsection{Protocol and scoring}

This section describes how the evaluation runs, how answers are scored in terms of criteria and formula, how the two evaluation modes apply, and how results are reported. All models in the main evaluation share a single tool library, organized into six families: \code{search}, \code{read}, \code{download}, \code{table\_operations}, \code{bash}, and \code{submit\_answer}. Each task allows 7,200 seconds and up to three submissions, using the same public inputs and cutoff-eligible data~\citep{prophetlab2026ile}; prompts and tool schemas are given in Appendix~\ref{app:agent-interface}.

Answers are scored against a common standard. All tasks use weighted criteria on a common 0--100 scale, and the rubric dimensions describe the requirements assessed within an answer: requested outputs (Result), source support and measurement basis (Evidence), economic interpretation (Financial logic), numerical operations (Calculation), consistency among linked quantities (Reconciliation), alignment of conclusions with the analysis and mandate (Decision consistency), and risks and sensitivity (Uncertainty). Definitions and examples are given in Appendix~\ref{app:dimension-definitions}. Scoring is carried out by two kinds of judges: numerical criteria are scored by deterministic checks, and semantic criteria are judged by GPT-6 Astra. Model-based evaluation can exhibit position, verbosity, and self-enhancement biases~\citep{zheng2023llmjudge}; Appendix~\ref{sec:judge-comparison} reports sensitivity to the choice of judge. Market tasks instead use a single 100-point Result criterion, computed from the declared price or ranking loss.

Given criterion weights $w_{ij}$ and earned fractions $g_{ij}\in[0,1]$, the score for task $i$ is
\begin{equation}
s_i = 100 \left( \frac{\sum_{j} w_{ij} g_{ij}}{\sum_{j} w_{ij}} \right),
\qquad
S = \frac{1}{N} \sum_{i=1}^{N} s_i,
\label{eq:score}
\end{equation}
where every task contributes equally to the overall score $S$.

The same scoring mechanism applies under two evaluation modes. \textbf{Strict evaluation} requires the answer to be successfully submitted through \code{submit\_answer}, which validates the deliverable and returns format errors for correction; under this mode, Step 5 Preview drops from 55.1 to 34.5 and Hunyuan 4 Preview from 60.4 to 47.7 (Table~\ref{tab:submission-recovery}). To distinguish financial reasoning quality from submission-format compliance, the default \textbf{answer-based evaluation} scores the final available answer using the original criteria, drawing from the accepted submission, the last submission payload, or the terminal assistant response. Tasks without a final answer receive zero credit. Scores under the default mode are those used for the main results below, and full strict results are reported in Appendix~\ref{app:submission-rescoring}.

Results are reported at two levels. Table~\ref{tab:results}(a) averages task-normalized dimension credit over applicable tasks, characterizing how models differ across scoring dimensions; Table~\ref{tab:results}(b) averages complete task scores within each professional category, characterizing how models perform across task types; and Figure~\ref{fig:diagnostics} summarizes both profiles together with overall scores, with aggregation and coverage detailed in Appendix~\ref{app:scope}.

\subsection{Performance across financial tasks}

Strong financial analysis does not necessarily translate into accurate forecasts. We examine whether models that excel at evidence-based research and financial reasoning also perform well on forward-looking prediction tasks. Astra leads overall at 68.1, followed by Opus at 66.9 (Table~\ref{tab:results}), yet their profiles of strength differ: Opus leads Result and Calculation, while Astra leads Financial logic, Reconciliation, Decision consistency, and Uncertainty, and Qwen Max leads Evidence (Table~\ref{tab:results}a). The difference concentrates in forecasting. Astra scores 78.9 on Research \& Evidence and 77.6 on Evaluate \& Decide, but only 52.5 on Forecast \& Model and 66.3 on Integrated Research, with KPI and cross-company forecasting at 41.1 and 49.6. Opus leads Forecast \& Model (52.6) and Monitor \& Update (64.2). These scores jointly point to a persistent weakness in estimating future drivers and peer effects. When predictive performance is taken as the yardstick, the ranking changes: DeepSeek V4.1 Flash leads at 40.3, ahead of Astra at 37.7 and Opus at 37.6, and DeepSeek V4 Pro leads the Market Trading category at 64.6 despite an overall score of 59.3 (Appendix~\ref{app:forecast-outcomes}). Overall research quality and predictive performance therefore capture different strengths.

\begin{figure}[!t]
\centering
\includegraphics[width=\linewidth]{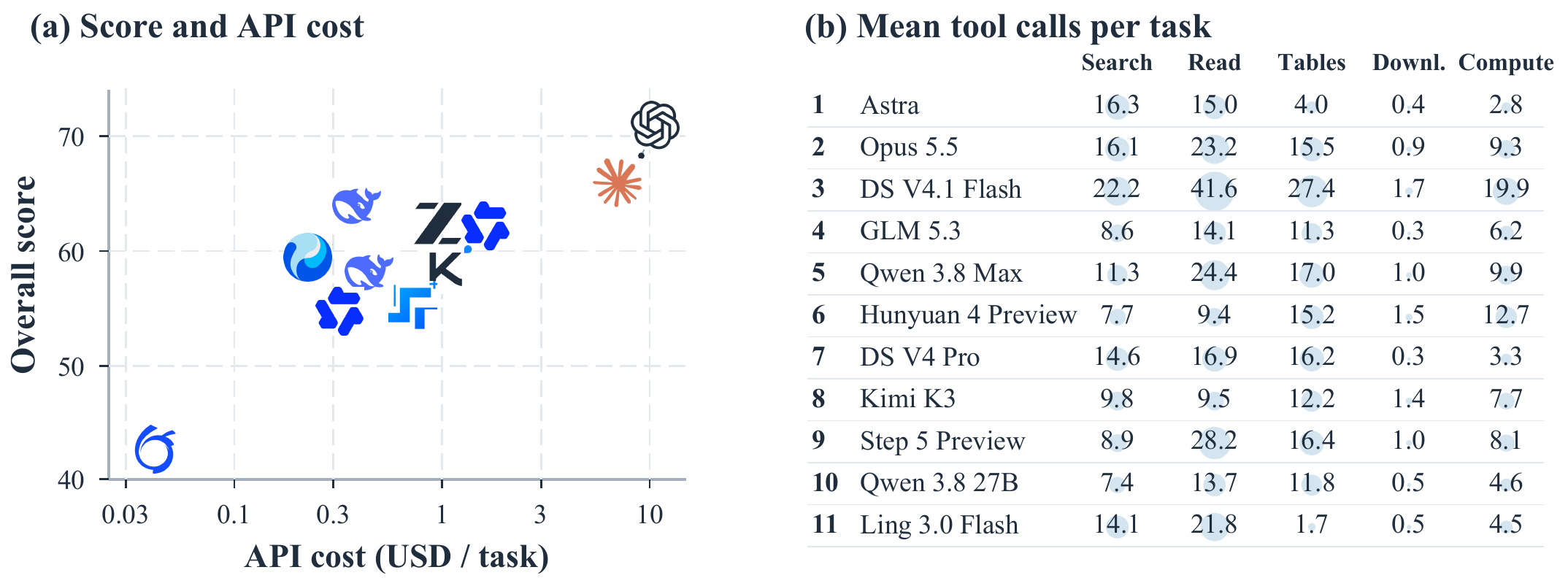}
\caption{\textbf{Score, cost, and tool use.} (a) Overall score versus API cost per task. (b) Mean tool calls, grouping table operations.}
\label{fig:resources}
\end{figure}

Beyond final scores, the demanding multi-step research workflows in ILE reveal substantial differences in models' tool-use patterns and agentic capabilities. DeepSeek V4.1 Flash averages 73.8 interaction rounds and 113.8 tool calls per task, making the most retrieval and computation calls, while the overall leader Astra makes fewer; Kimi K3 averages 26.7 rounds and DeepSeek V4 Pro 34.1, both above the 10.4 and 21.3 reported by FrontierFinance~\citep{zhang2026frontierfinance}, and Finance Agent Benchmark reports 13.5 rounds for o3~\citep{2025_FinanceAgentBenchmark} (Appendix~\ref{app:resources}). Opus and Qwen Max have similar read and computation counts yet differ by 4.8 points. On cost, DeepSeek V4.1 Flash scores 63.0 at USD 0.36 per task, whereas Opus gains 3.9 points at USD 7.96, roughly 22 times the cost (Figure~\ref{fig:resources}). These results highlight the agentic demands of financial research, where effective performance requires not merely extensive tool use, but the ability to coordinate evidence retrieval, computation, and financial reasoning across multi-step workflows.

\subsection{From fundamental forecasts to stock returns}\label{sec:trading}
\noindent\begin{minipage}[t]{.50\linewidth}
\vspace{0pt}\raggedright
To test whether forecasts translate into realized returns, we construct trades on a retrospective sample of twenty assignments covering eleven companies and thirteen disclosures. Forecasts above or below archived pre-cutoff analyst estimates generate long or short signals, while conflicting votes and missing forecasts hold cash. Capital is allocated equally across opportunities; entry is the first close after the cutoff and exit is the first trading-day close after disclosure, with costs of 10 basis points per leg plus 3\% annualized short borrowing.\par\medskip
Hunyuan leads this sample at 8.50\%, followed by Flash at 7.98\% and Astra at 7.18\%, versus 1.18\% for always long. Kimi's 7.15\% exceeds Opus's 6.39\% despite its lower research score. Returns depend on forecast coverage, direction relative to expectations, and the subsequent stock move; the capability measured by research scores and the capability reflected in trading returns do not coincide, and the former does not guarantee the latter (Appendix~\ref{app:trading}).
\end{minipage}\hfill
\begin{minipage}[t]{.48\linewidth}
\vspace{0pt}
\raggedright\small
\setlength{\abovecaptionskip}{0pt}
\setlength{\belowcaptionskip}{5pt}
\setlength{\tabcolsep}{2.5pt}
\renewcommand{\arraystretch}{1.02}
\captionof{table}[Forecast-driven stock returns]{\raggedright\textbf{Forecast-driven stock returns, March--August 2026.} Return and maximum drawdown (MDD) are reported in percent, and excess return relative to always long in percentage points.\par}
\label{tab:trading}
\begin{tabularx}{\linewidth}{@{}Yrrr@{}}
\toprule
System & Return & Excess & MDD \\
\midrule
Hunyuan 4 Preview & $+8.50$ & $+7.32$ & 2.72 \\
DeepSeek V4.1 Flash & $+7.98$ & $+6.80$ & 3.65 \\
GPT-6 Astra & $+7.18$ & $+6.00$ & 1.50 \\
Kimi K3 & $+7.15$ & $+5.97$ & 4.02 \\
Claude Opus 5.5 & $+6.39$ & $+5.21$ & 2.94 \\
GLM 5.3 & $+6.19$ & $+5.01$ & 3.53 \\
Qwen 3.8 27B & $+5.20$ & $+4.02$ & 1.90 \\
DeepSeek V4 Pro & $+3.64$ & $+2.46$ & 2.03 \\
Ling 3.0 Flash & $+2.51$ & $+1.33$ & 2.64 \\
Qwen 3.8 Max & $+1.67$ & $+0.48$ & 2.05 \\
\midrule
Always long & $+1.18$ & $+0.00$ & 7.63 \\
Always short & $-2.27$ & $-3.45$ & 8.86 \\
\bottomrule
\end{tabularx}
\end{minipage}
\par\medskip

\subsection{Public-web research and database coverage}\label{sec:codex-main}
Further, we hold the shared task set, scoring standard, and cutoff fixed and vary only the evidence source, in order to examine whether comparable scores imply comparable compliance. Astra scores 68.1 on our database and 68.3 on public-web Codex, while Qwen Max and Hunyuan score higher on our database (Appendix~\ref{app:codex}). Comparable scores, however, do not imply a comparable boundary: 57\% of Astra's public-web trajectories contain post-cutoff sources, including the target quarter's realized Microsoft revenue and Tesla deliveries, despite a dated query (Figure~\ref{fig:research-cases}). These trajectories draw on evidence available only after the cutoff, and their scores cannot be regarded as obtained within the information boundary. Comparable scores therefore establish comparable answer quality only; whether the evidence precedes the cutoff cannot be read off the score and is guaranteed only by the evidence source itself.

\Needspace*{8\baselineskip}
\section{Where Financial Research Breaks Down}
Our case analyses reveal three recurring challenges in financial research: tracing cash flows to funding needs while distinguishing sustainable improvements from temporary movements, evaluating the predictive validity of financial assumptions, and ensuring that assumptions and outputs form a consistent financial model. We illustrate these challenges through representative tasks, then examine submission reliability and the role of expert rubric refinement.
\subsection{A complete case: from cash flows to a funding judgment}\label{sec:case-aep-main}
\textbf{Question, dependency, and cash-flow quality.} AEP N27 asks how funding needs and fuel-cost recovery changed between the first halves of 2025 and 2026. Its report must connect disclosed cash flows to external financing needs and explain the quality of the change (Figure~\ref{fig:research-cases}, right). The first breakdown lies in the dependency itself. Operating cash flow rises from USD 2,671 million to 3,421 million, while investing cash use rises from 5,347 million to 6,602 million; the pre-financing shortfall, investing use less operating inflow, therefore widens from 2,676 million to 3,181 million, an increase of 505 million. Construction spending grows by 1,586 million, exceeding the 1,255 million increase in net investing use because other investment items partly offset it. The second breakdown lies in explaining cash-flow quality. Net income falls by 441 million and non-cash adjustments rise by 516 million, together explaining only 75 million of the 750 million operating-cash improvement, with the remainder coming mainly from other noncurrent-liability movements that contribute 540 million, driven by customer deposits and rate-rider refund timing. The net deferred-fuel adjustment improves by 72 million but does not identify gross fuel collections. Without this distinction, operating improvement is easily misread as sustainable cash generation.

\textbf{What separates systems.} The eleven evaluated systems all recover the eight reported inputs and reconcile both cash-flow identities, yet only five compute both funding shortfalls, three distinguish net fuel adjustments from gross collections, and five reach the supported funding conclusion. Recovering inputs and completing calculations therefore does not amount to understanding the dependencies; an answer must separate operating improvement from temporary cash movements and combine that assessment with capital needs. Appendix~\ref{app:case-aep} gives the full decomposition.

\subsection{Calculation, forecasting, and joint consistency}
\textbf{Predictive validity cannot be established by arithmetic.} In Tesla O80, three systems earn every supporting-analysis point but no forecast-outcome credit. Their units, revenue bridges, and recommendations are coherent, yet deliveries, revenue, and margin all miss their realized targets. Historical observations can support several plausible assumptions about future demand, pricing, and cost, and arithmetic propagates whichever assumptions are chosen without establishing their predictive validity (Appendix~\ref{app:capability-cases}).

\begin{samepage}
\textbf{Assumptions and outputs must satisfy jointly testable consistency constraints.} At the accounting level, Home BancShares N41 requires expense, payment, and payable forecasts to describe a single accounting state, and any forecast that departs from that state fails. At the level of argument, GE Aerospace N06 requires a position limit to follow from a supported equity-downside assumption; Qwen Max earns full formula credit yet offers no explanation of its protection boundary, showing that a correct formula does not amount to a valid argument. Disclosed obligations, accounting reserves, and equity losses measure different quantities, and combining them into a downside estimate requires an explicit financial argument. Both kinds of tasks test whether assumptions and outputs form a valid financial model (Appendix~\ref{app:case-ge}).\par
\end{samepage}

\subsection{Submission Reliability and Task Completion}
Submission failures reveal a distinct limitation of financial research agents: producing useful analysis does not guarantee successful completion of the required deliverable. Hunyuan stops without submitting on 17 trajectories, and Opus has three rejected submissions on each of five. Such failures prevent answer quality from translating into task completion, and Table~\ref{tab:submission-recovery} quantifies this gap, with full strict results in Appendix~\ref{app:submission-rescoring}.

\subsection{What expert refinement improves}\label{sec:refinement}
\noindent\begin{minipage}[t]{.42\linewidth}
\vspace{0pt}\raggedright
If scoring criteria remain at the level of general financial requirements, errors cannot be traced to a specific step and therefore cannot be attributed. Expert review serves to break such requirements into distinguishable criteria, so that each kind of error has a corresponding test. Microsoft O58 separates additive reporting segments from overlapping Cloud revenue, making double counting a testable scope error. JPMorgan O26 fixes a dated prior before peer disclosures, so that an update must be attributable to new evidence rather than an unstated baseline. 
\end{minipage}\hfill
\begin{minipage}[t]{.56\linewidth}
\vspace{0pt}\small
\setlength{\abovecaptionskip}{0pt}
\setlength{\belowcaptionskip}{5pt}
\setlength{\tabcolsep}{2.5pt}
\renewcommand{\arraystretch}{1.12}
\captionof{table}[Score drops under strict submission requirements]{\raggedright\textbf{Score drops under strict submission requirements.} Only affected systems are shown; change is strict minus answer-based score.}
\label{tab:submission-recovery}
\begin{tabularx}{\linewidth}{@{}Yrrr@{}}
\toprule
System & \shortstack{Answer-\\based} & Strict & Change \\
\midrule
Step 5 Preview & 55.12 & 34.46 & $-20.66$ \\
Hunyuan 4 Preview & 60.39 & 47.73 & $-12.66$ \\
DeepSeek V4.1 Flash & 62.99 & 53.03 & $-9.96$ \\
Claude Opus 5.5 & 66.87 & 62.69 & $-4.18$ \\
Ling 3.0 Flash & 42.59 & 40.93 & $-1.66$ \\
\bottomrule
\end{tabularx}
\end{minipage}
\par\medskip

AWS O15 jointly scores interval accuracy and uncertainty, preventing a narrow miss from earning width-only credit. Each of these requirements links to a source passage, calculation, or realized outcome, so the revised rubrics can locate an error in a specific financial dependency. Matched revision scores are reported in Appendix~\ref{app:curation}.

\section{Limitations}\label{sec:limitations}



Although ILE restricts look-ahead bias during retrieval through its versioned database, frontier models may already have implicitly memorized some historical outcomes during pre-training~\citep{glasserman2023lookahead}; fully disentangling internal knowledge from external retrieval will require ``counterfactual'' or ``cold-start'' experiments.
In addition, some ILE tasks are settled against realized disclosures or market prices, and correct intermediate reasoning does not guarantee an accurate forecast: exogenous market noise can lead otherwise sound analyses to numerically deviate from realized values, so evaluating agents solely on outcome accuracy would underestimate research quality.

In future work, we will pursue two directions. First, we will build a counterfactual and synthetic financial-scenario engine that perturbs real financial statements and constructs virtual peer companies, forcing agents to reason solely from retrieved evidence and thereby isolating parametric memory at its source. Second, we will restructure tasks and evaluation: decomposing long expert-authored assignments into composable micro-task graphs, combined with difficulty-weighted scoring and a multi-model consensus judging protocol; and for forecast-type tasks, we will also refine the evaluation to better distinguish reasoning failures from stochastic fluctuations, advancing ILE toward a living benchmark that can expand continuously with new market disclosures.

\section{Conclusion}

We introduce Investor's Last Exam (ILE), a benchmark for professional equity research that combines a versioned financial database, a professional task taxonomy, and an expert-led construction framework. ILE enforces decision-time information boundaries across retrieval and computation over 3.43 million document versions, while financial dependency graphs link scoring criteria to source evidence, calculations, and assumptions. Experiments on eleven systems show that strong supporting analysis does not consistently translate into accurate forecasts and well-supported financial conclusions: GPT-6 Astra leads overall at 68.1, but scores 52.5 on Forecast \& Model. Forecast-outcome scores and simulated investment returns further produce different model rankings. Dependency-level analysis locates errors in driver assumptions, accounting scope, and the financial interpretation of calculated quantities. By making the evidence--assumption--judgment chain the unit of evaluation, ILE connects research quality to realized outcomes and provides a traceable, diagnostic foundation for measuring and improving financial agents' ability to turn evidence into reliable forecasts and defensible investment decisions.
\clearpage
\begingroup\small\raggedright
\bibliographystyle{plainnat}
\bibliography{references}
\endgroup

\clearpage
\appendix
\section{Related Work}\label{sec:related}
\paragraph{Financial questions and research workflows.}
FinQA evaluates numerical reasoning over financial documents~\citep{2021_FinQA}; TAT-QA combines tables and associated text~\citep{zhu2021tatqa}, and ConvFinQA extends numerical reasoning to conversational chains~\citep{chen2022convfinqa}. FinBen broadens evaluation across financial tasks, including forecasting and decision-making~\citep{xie2024finben}. FinanceBench pairs open-book questions with answers and evidence~\citep{2023_FinanceBench}. Finance Agent Benchmark and its v2 extension evaluate expert-authored analyst assignments with retrieval tools~\citep{2025_FinanceAgentBenchmark,vals2026fabv2}. FinSearchComp studies professional financial search and reasoning~\citep{2025_FinSearchComp}; DeepResearch Bench evaluates longer research reports and citations beyond finance~\citep{2025_DeepResearchBench}. \ILE\ organizes tasks by their financial deliverable and couples research with a task-specific information boundary.

\paragraph{Professional deliverables and expert rubrics.}
GDPval evaluates expert-created work products across occupations~\citep{patwardhan2025gdpval}. APEX-Agents extends professional evaluation to long-horizon, cross-application workflows, including investment banking~\citep{vidgen2026apexagents}; BankerToolBench evaluates end-to-end investment-banking workflows through financial models, presentations, and reports graded against expert-defined criteria~\citep{lau2026bankertoolbench}. These benchmarks motivate evaluation through professional deliverables. \ILE\ focuses on buy-side equity research under version-level information cutoffs and additionally evaluates designated forecasts against realized outcomes. Beyond finance, PaperBench uses expert-developed hierarchical rubrics for research replication~\citep{starace2025paperbench}, while LLM-Rubric studies multidimensional evaluation and calibration~\citep{hashemi2024llmrubric}. RubricBench documents a gap between human-authored and model-generated criteria~\citep{zhou2026rubricbench}, motivating explicit validation of the criteria themselves.

\paragraph{Evidence-linked assessment.}
BigFinanceBench uses weighted rubrics for derivation steps~\citep{wang2026bigfinancebench}; Samaya's FrontierFinance uses expert-authored, source-attributed rubrics across investor workflows~\citep{zhang2026frontierfinance}. The distinct FrontierFinance benchmark from Krumdick et al. emphasizes long-horizon computer-use financial modeling and includes human evaluation~\citep{2026_FrontierFinance_Kensho}. Classical data-provenance research distinguishes source locations from the input dependencies that explain an output~\citep{buneman2001provenance}. ALCE evaluates citation-supported generation~\citep{gao2023alce}, and VeriScore assesses verifiable claims in long-form responses~\citep{song2024veriscore}. \ILE's financial dependency graph connects the proposition being graded to the calculation, assumption, and source version that support it. Experts can inspect and edit these dependencies while allowing solvers to develop their own analytical route.

\paragraph{Forecasting and historical information environments.}
ForecastBench evaluates future-event forecasts~\citep{karger2025forecastbench}; Bench to the Future supplies an offline pastcasting corpus~\citep{2025_BenchToTheFuture}. FinDeepForecast constructs live financial forecasts over a timestamped database~\citep{li2026findeepforecast}, and FinBench focuses on time-gated calibration~\citep{ghosh2026finbench}. Agentic Time Machine approximates historical web access~\citep{2026_AgenticTimeMachine}. FinanceHarness/FinanceGym combines a self-collected, publication-dated web corpus with point-in-time search, an entity graph for task generation, expert validation, and pre-/post-cutoff criteria~\citep{xiao2026financeharness}. \ILE\ integrates versioned financial data, professional equity-research deliverables, and an expert-authored dependency-level construction workflow. Samaya's FrontierFinance instead uses dated queries and applies a retrieval cutoff to its in-house index while excluding predictive queries~\citep{zhang2026frontierfinance}. StockBench measures trading-agent performance~\citep{2025_StockBench}. Research on data vintages and look-ahead bias examines how temporal data access and pretrained knowledge affect forecasting evaluations~\citep{2001_Stark_Croushore_RealTimeForecasting,2026_SearchDateLeakage_ACL,2026_LookAheadBench,2026_ExplicitBiasConsideration,2025_Pitfalls_Forecasters,glasserman2023lookahead}. LiveBench reduces contamination risk through regularly refreshed questions and objective scoring~\citep{white2025livebench}.

\section{Database Inventory}\label{app:inventory}
Table~\ref{tab:inventory} summarizes the financial document collection. The nine source categories contain 3,430,336 document versions, counting revisions separately, and occupy 1.52\,TB. Structured market data are provided separately. Each task accesses the source versions available at its cutoff within its authorized datasets.
\begin{table}[ht]
\centering\small
\caption{Document-version counts by source category, 15 October 2026.}
\label{tab:inventory}
\begin{tabular}{@{}lr@{}}
\toprule
Data type & Document versions \\
\midrule
Financial reports & 253,055 \\
Financial statements & 876,903 \\
Earnings calls & 136,651 \\
Investor materials & 49,601 \\
Issuer events & 356,574 \\
Analyst estimates & 425,380 \\
Research reports & 199,123 \\
Institutional holdings & 9,379 \\
News & 1,123,670 \\
\midrule
Total & 3,430,336 \\
\bottomrule
\end{tabular}
\end{table}

The collection covers US, mainland Chinese, Hong Kong, and other markets. Formats include structured tables, text, PDFs, images, and office documents.

\subsection{Versioned data architecture}
The database links issuers, securities, financial records, and document versions within a shared catalogue. Immutable source files and their derived representations are linked to reporting periods, publication times, and supporting time evidence.

\subsection{Cutoff enforcement}\label{app:environment-contract}
Each task specifies a cutoff $t_c$ and an authorized access limit $t_{\mathrm{auth}}$. The service uses $C=\min(t_c,t_{\mathrm{auth}})$ and releases only authorized versions satisfying $a(v)<C$, where $a(v)$ is the version's availability time. Source timestamps retain their timezone and precision; date-only records become eligible after that date has ended. By default, discovery selects the latest collected version available before the cutoff, before matching the query. An explicit history mode exposes other eligible versions.

Search, direct reads, downloads, and cached results enforce the same authorization and availability rules. Structured files spanning publication dates are filtered before release to the task workspace.

Downloads and local computation share the task's isolated workspace. Reference analyses, rubrics, and later outcomes are held in a separate evaluation environment. Forecasts are matched to realized values by metric, period, and unit; consensus estimates serve as research inputs.

\section{Professional Task Taxonomy}\label{app:taxonomy}
Table~\ref{tab:taxonomy-full} links each category to its professional deliverable and judgment.

\begin{table}[!ht]
\centering\small
\caption{Professional deliverables and judgments in each task category. The first five categories contain twenty subtypes; integrated research combines them, and Market Trading evaluates realized market outcomes.}
\label{tab:taxonomy-full}
\begin{tabularx}{\linewidth}{@{}>{\raggedright\arraybackslash}p{.19\linewidth} r Y Y@{}}
\toprule
Category / format & $n$ & Analyst deliverable & Professional judgment \\
\midrule
Discover \& Frame & 10 & Screened opportunities, catalyst calendars, a thesis, or a driver map. & Identify economically material questions and testable investment hypotheses. \\
Research \& Evidence & 13 & Reconciled facts, historical series, unit economics, or supply-chain relationships. & Resolve definitions and accounting scope; establish which evidence supports a claim. \\
Forecast \& Model & 13 & Operating forecasts, earnings bridges, financial statements, or peer read-throughs. & Link drivers to financial outputs; distinguish observed signals from assumptions. \\
Evaluate \& Decide & 14 & Valuation, implied expectations, surprise probabilities, or an action under stated payoffs. & Compare consistent measures and translate uncertainty into an investment decision. \\
Monitor \& Update & 17 & Material-event assessments, revised forecasts, thesis tests, or position adjustments. & Identify incremental information and revise a prior without double-counting. \\
\midrule
Integrated Research & 11 & A research report connecting forecasts, investment stance, risks, and thesis conditions. & Maintain consistency from evidence through model to recommendation. \\
Market Trading & 8 & A price forecast or a return/volatility ranking for a defined horizon. & Anticipate realized market outcomes under the same information boundary. \\
\bottomrule
\end{tabularx}
\end{table}

Table~\ref{tab:subtypes} describes the 20 professional subtypes, organized into five categories with four subtypes each. Integrated Research combines these analytical functions in a single assignment. Market Trading adds three outcome-based subtypes. Each task belongs to one group according to its principal objective; evidence requirements and analytical techniques can overlap.
\begin{table}[ht]
\centering\small
\caption{Twenty analyst-work subtypes, integrated research, and three market-forecasting subtypes. Descriptions specify the intended analytical output.}
\label{tab:subtypes}
\begin{tabularx}{\linewidth}{@{}>{\raggedright\arraybackslash}p{.39\linewidth} Y@{}}
\toprule
Professional subtype & Analytical output \\
\midrule
\multicolumn{2}{@{}l}{\textbf{Discover \& Frame}} \\
1.1 Screening \& ranking & A ranked opportunity set under explicit investment criteria. \\
1.2 Catalyst calendars & Dated events and their possible effects on the investment thesis. \\
1.3 Thesis evidence & An investment hypothesis with supporting and opposing evidence. \\
1.4 Drivers \& sensitivity & Material operating or financial drivers and comparable sensitivities. \\
\midrule
\multicolumn{2}{@{}l}{\textbf{Research \& Evidence}} \\
2.1 Information extraction & Source-grounded facts with consistent definitions, periods, and units. \\
2.2 Historical reconstruction & A comparable time series with reporting changes reconciled. \\
2.3 Unit economics & Revenue, cost, and contribution per product, customer, or operating unit. \\
2.4 Supply-chain mapping & Economic relationships, exposures, and transmission channels. \\
\midrule
\multicolumn{2}{@{}l}{\textbf{Forecast \& Model}} \\
3.1 KPI forecasting & Operating-metric forecasts justified by observable drivers. \\
3.2 Revenue, margin \& EPS & Earnings forecasts with internally consistent financial bridges. \\
3.3 Three-statement modelling & Linked income, balance-sheet, and cash-flow projections. \\
3.4 Cross-company forecasting & A justified read-through from one company's signals to another. \\
\midrule
\multicolumn{2}{@{}l}{\textbf{Evaluate \& Decide}} \\
4.1 Valuation & A value estimate with explicit economic assumptions and sensitivities. \\
4.2 Reverse valuation & The operating expectations implied by a market price. \\
4.3 Consensus \& earnings surprises & A like-for-like comparison with consensus and surprise estimates. \\
4.4 Probabilistic decisions & An action justified by probabilities, payoffs, and constraints. \\
\midrule
\multicolumn{2}{@{}l}{\textbf{Monitor \& Update}} \\
5.1 Material events & An assessment of new information's financial materiality. \\
5.2 Forecast updates & A revised forecast isolating the contribution of new evidence. \\
5.3 Thesis testing & Evidence-based confirmation, revision, or rejection of a thesis. \\
5.4 Position management & A position adjustment consistent with the mandate and revised view. \\
\midrule
\textbf{Integrated Research} & A coherent report linking evidence, forecasts, valuation or stance, risks, and thesis conditions. \\
\midrule
\textbf{Market Trading} & Outcome-based predictions at a defined future horizon. \\
7.1 Price forecasts & A single stock's future closing price. \\
7.2 Return rankings & Securities ordered by their realized forward returns. \\
7.3 Volatility rankings & Securities ordered by their realized forward volatility. \\
\bottomrule
\end{tabularx}
\end{table}

\subsection{Financial dependencies across assignments}
\textbf{Comparable measurements must be reconstructed before they can be modeled.} Tesla task O70 asks for annual automotive revenue over FY2016--2025, its sales, regulatory-credit, and leasing components, and a bridge from income-statement to segment presentation. The solver must identify changes in reporting definitions and document their effect on comparability. Home BancShares task N41 requires forecasts of interest expense, cash interest paid, and closing interest payable that obey one accounting identity. These tasks assess whether individually plausible numbers form a valid financial representation.

\textbf{New evidence must justify a specific change in a forecast.} JPMorgan task O26 requires a dated prior, a capital-markets peer signal, and an explicit coefficient translating that signal into investment-banking fees and Markets revenue. Unaffected lines must retain their prior values, and the changes must reconcile to the updated totals. AWS task O15 similarly requires an itemized forecast-revision bridge and reasons for leaving other drivers unchanged. The required judgment is the magnitude and scope of the update supported by each disclosure.

\textbf{An investment decision must remain consistent with its assumptions and reversal conditions.} Apple task O09 combines an earnings forecast with an investment stance and material risks. JPMorgan task O32 asks which earnings drivers are durable and which developments would change the stance. GE Aerospace task N06 requires a loss-constrained equity weight supported by an explicit downside assumption, while distinguishing supported obligations, accounting reserves, and guarantee exposure. These assignments test whether the agent can turn financial analysis into a decision whose assumptions and conditions can be examined.

\section{A Worked Financial Dependency}\label{app:example}
JPMorgan task O26 asks how a capital-markets peer disclosure changes a previously formed forecast. The solver first fixes a Q2 2026 prior using information available by March 31, 2026, 16:00 ET, then incorporates a peer signal available by the task's May 14 cutoff. Let $I_0$ and $M_0$ denote prior investment-banking fees and Markets revenue, $g_I$ and $g_M$ the peer's disclosed growth rates, and $\beta$ the chosen read-through coefficient. The affected-line updates are
\begin{equation}
\Delta I=\beta g_I I_0,\qquad \Delta M=\beta g_M M_0,
\end{equation}
with updated totals obtained by adding these increments to the prior and holding other lines fixed. The source and period of each peer signal, the dated prior, the choice of $\beta$, and the reconciliation to CIB and total revenue are distinct verification targets. For example, Astra's revised answer uses $I_0=2{,}714.04$, $g_I=0.12$, and $\beta=0.5$, producing $\Delta I=162.8424$ million. The equation makes arithmetic reproducible; source evidence and the coefficient's justification establish whether the update is financially supported.

\section{Rubric Dimensions and Score Aggregation}\label{app:scope}
\subsection{Dimension definitions}\label{app:dimension-definitions}
Rubric dimensions classify the financial requirement evaluated by an individual criterion. Each criterion carries one dimension label, while an assignment can contain criteria from several dimensions. Professional task categories instead identify the assignment's principal deliverable (Appendix~\ref{app:taxonomy}). Table~\ref{tab:rubric-dimensions} defines the dimensions using examples from the benchmark's rubrics.

\begin{table}[!ht]
\centering\small
\setlength{\tabcolsep}{4pt}
\renewcommand{\arraystretch}{1.12}
\caption{\textbf{Rubric dimensions: assessed requirements and examples.} Dimensions describe requirements within answers; professional task categories describe whole assignments.}
\label{tab:rubric-dimensions}
\begin{tabularx}{\linewidth}{@{}>{\raggedright\arraybackslash}p{.16\linewidth}YY@{}}
\toprule
Dimension & Assessed requirement & Example from a task rubric \\
\midrule
Result & Accuracy or completeness of the requested numerical targets, rankings, or other specified outputs. & AWS O15: the revenue point forecast is scored against realized segment net sales. \\
\addlinespace[3pt]
Evidence & Support from eligible sources using the required entity, period, unit, and reporting basis. & Screening O01: growth inputs use each issuer's as-reported rate and latest completed fiscal year. \\
\addlinespace[3pt]
Financial logic & Financial definitions, economic mechanisms, and assumptions that justify an inference or forecast. & AWS O15: each forecast adjustment has a magnitude justified through demand, capacity, pricing, or timing. \\
\addlinespace[3pt]
Calculation & Numerical execution of a stated transformation or formula using the specified inputs and units. & AEP N27: the pre-financing shortfall equals investing cash use less operating cash flow. \\
\addlinespace[3pt]
Reconciliation & Agreement among linked quantities under accounting identities, component totals, or financial bridges. & AEP N27: net cash flows reconcile to the cash increase, and opening cash plus that increase equals closing cash. \\
\addlinespace[3pt]
Decision consistency & Agreement of conclusions, rankings, or actions with the analysis, stated rules, and investment mandate. & Screening O01: the written shortlist matches the qualification flags and ranking in the screen. \\
\addlinespace[3pt]
Uncertainty & Quantification and explanation of uncertainty through intervals, probabilities, sensitivity, or conditions that change the conclusion. & Screening O01: using constant-currency growth would change Alphabet's qualification, making the shortlist sensitive to the reporting basis. \\
\bottomrule
\end{tabularx}
\end{table}

The distinction between Evidence, Calculation, and Reconciliation follows the object being checked. In O01, the Evidence criteria establish whether growth inputs use the required reporting basis and fiscal period. In N27, the Calculation criteria derive a funding shortfall from disclosed cash flows, while Reconciliation criteria check cash-flow and cash-balance identities across linked figures. Financial logic evaluates the interpretation of these relationships; Decision consistency evaluates whether the resulting conclusion follows from the analysis.

\subsection{Score aggregation}
\paragraph{Overall and professional-group scores.}
Equation~\ref{eq:score} first normalizes criterion credit within each task, then weights tasks equally. Task-category scores in Table~\ref{tab:results}(b) and Figure~\ref{fig:diagnostics}(c) average complete task scores within each category. The benchmark contains 78 company-research assignments and eight market-forecasting assignments; both contribute to the same overall score. A market task's deterministic outcome loss is mapped to a single 100-point Result criterion.

\paragraph{Assessment dimensions.}
Table~\ref{tab:results}(a), Figure~\ref{fig:diagnostics}(b), and Figure~\ref{fig:profiles-additional} use task-normalized dimension credit. Let $\mathcal{T}_d$ contain tasks with at least one criterion in dimension $d$. We compute
\begin{equation}
M_d=\frac{1}{|\mathcal{T}_d|}\sum_{i\in\mathcal{T}_d}
100\frac{\sum_{j\in d}w_{ij}g_{ij}}{\sum_{j\in d}w_{ij}}.
\end{equation}
Result has coverage across all 86 tasks. Evidence, Financial logic, Calculation, Reconciliation, Decision consistency, and Uncertainty have coverage on 66, 61, 59, 47, 62, and 43 tasks, respectively. A task without a criterion in dimension $d$ is excluded from $\mathcal{T}_d$; an applicable task without a final answer contributes zero. Dimension means summarize different task sets and are computed separately from the overall score.

\paragraph{Answer extraction and submission handling.}
The default evaluation uses an accepted answer when available. For a failed submission, it grades the last emitted \texttt{submit\_answer} payload; when the tool is never called, it uses the terminal assistant answer. Extraction preserves the emitted financial content. Each answer is evaluated with the original task-specific criteria, weights, and judge. Runs without an identifiable final answer retain zero. Strict evaluation requires a valid, accepted submission, so both omitted calls and rejected submissions receive zero. Table~\ref{tab:submission-audit} records these failures; Table~\ref{tab:strict-results} reports their effect on the full dimension and category breakdown.

\begin{table}[htbp]
\centering\footnotesize
\setlength{\tabcolsep}{3pt}
\renewcommand{\arraystretch}{1.12}
\caption{\textbf{Submission omissions and strict scores.} Report and Market columns count tasks with no \texttt{submit\_answer} call among 37 report and eight market tasks. Strict zeros cover omitted and rejected submissions across all tasks.}
\label{tab:submission-audit}
\begin{tabularx}{\linewidth}{@{}Yrrrrr@{}}
\toprule
Model & \shortstack{Report\\$n=37$} & \shortstack{Market\\$n=8$} & \shortstack{Combined\\$n=45$} & \shortstack{Strict\\zeros} & \shortstack{Strict\\score} \\
\midrule
Hunyuan 4 Preview & 16 (43.2\%) & 1 (12.5\%) & 17 (37.8\%) & 17 & 47.7 \\
DeepSeek V4.1 Flash & 14 (37.8\%) & 0 & 14 (31.1\%) & 14 & 53.0 \\
Ling 3.0 Flash & 5 (13.5\%) & 2 (25.0\%) & 7 (15.6\%) & 10 & 40.9 \\
Step 5 Preview & 1 (2.7\%) & 0 & 1 (2.2\%) & 32 & 34.5 \\
GPT-6 Astra & 0 & 0 & 0 & 0 & 68.1 \\
Claude Opus 5.5 & 0 & 0 & 0 & 5 & 62.7 \\
DeepSeek V4 Pro & 0 & 0 & 0 & 0 & 59.3 \\
Qwen 3.8 Max & 0 & 0 & 0 & 0 & 62.1 \\
Qwen 3.8 27B & 0 & 0 & 0 & 0 & 54.6 \\
GLM 5.3 & 0 & 0 & 0 & 0 & 62.3 \\
Kimi K3 & 0 & 0 & 0 & 0 & 58.2 \\
\bottomrule
\end{tabularx}
\end{table}

\subsection{Full results with strict submission requirements}\label{app:submission-rescoring}
Table~\ref{tab:strict-results} reports strict scores by dimension and task category. Omitted or rejected submissions contribute zero to each applicable mean.

\begin{table}[!ht]
\centering\footnotesize
\setlength{\tabcolsep}{2.0pt}
\renewcommand{\arraystretch}{1.10}
\caption{\textbf{Strict submission evaluation.} An accepted \texttt{submit\_answer} call is required; submission failures receive zero. Both panels use the same tasks, criteria, and aggregation as Table~\ref{tab:results}. Bold marks the highest displayed score.}
\label{tab:strict-results}
\begin{tabular*}{\linewidth}{@{\extracolsep{\fill}}lrrrrrrrr@{}}
\toprule
\multicolumn{9}{@{}l}{\textbf{(a) Scoring dimensions: quality within answers}} \\
System & Overall & Result & Evidence & Logic & Calc. & Recon. & Decision & Uncert. \\
\midrule
GPT-6 Astra & \textbf{68.1} & \textbf{59.9} & 79.0 & \textbf{77.8} & \textbf{89.4} & \textbf{86.8} & \textbf{84.4} & \textbf{83.6} \\
Claude Opus 5.5 & 62.7 & 57.8 & 71.2 & 69.7 & 81.6 & 71.3 & 71.4 & 73.2 \\
GLM 5.3 & 62.3 & 56.9 & 79.0 & 71.3 & 82.1 & 84.3 & 74.1 & 76.3 \\
Qwen 3.8 Max & 62.1 & 55.8 & \textbf{81.1} & 74.6 & 86.5 & 80.3 & 70.0 & 71.9 \\
DeepSeek V4 Pro & 59.3 & 54.3 & 70.1 & 69.2 & 81.5 & 82.1 & 71.7 & 75.1 \\
Kimi K3 & 58.2 & 51.5 & 69.5 & 67.4 & 82.8 & 76.7 & 73.7 & 76.7 \\
Qwen 3.8 27B & 54.6 & 47.3 & 68.6 & 63.4 & 78.5 & 74.9 & 69.1 & 69.1 \\
DeepSeek V4.1 Flash & 53.0 & 48.5 & 69.5 & 64.4 & 70.5 & 69.5 & 67.4 & 73.4 \\
Hunyuan 4 Preview & 47.7 & 42.2 & 63.0 & 61.1 & 72.0 & 69.5 & 58.8 & 72.5 \\
Ling 3.0 Flash & 40.9 & 41.7 & 46.2 & 42.9 & 58.6 & 51.1 & 54.1 & 37.4 \\
Step 5 Preview & 34.5 & 36.7 & 31.3 & 28.8 & 34.6 & 31.5 & 26.6 & 33.1 \\
\bottomrule
\end{tabular*}
\par\vspace{7pt}
\begin{tabular*}{\linewidth}{@{\extracolsep{\fill}}lrrrrrrr@{}}
\toprule
\multicolumn{8}{@{}l}{\textbf{(b) Task categories: performance across assignments}} \\
System & \shortstack{Discover\\\& Frame} & \shortstack{Research\\\& Evidence} & \shortstack{Forecast\\\& Model} & \shortstack{Evaluate\\\& Decide} & \shortstack{Monitor\\\& Update} & \shortstack{Integrated\\Research} & \shortstack{Market\\Trading} \\
\midrule
GPT-6 Astra & \textbf{79.5} & \textbf{78.9} & \textbf{52.5} & \textbf{77.6} & 63.9 & \textbf{66.3} & 57.0 \\
Claude Opus 5.5 & 78.0 & 68.4 & 48.9 & 70.1 & \textbf{64.2} & 51.2 & 56.3 \\
GLM 5.3 & 77.3 & 67.1 & 46.6 & 70.7 & 62.2 & 56.9 & 53.8 \\
Qwen 3.8 Max & 72.3 & 74.4 & 49.9 & 66.0 & 57.8 & 57.8 & 57.1 \\
DeepSeek V4 Pro & 70.2 & 65.3 & 45.6 & 66.9 & 52.5 & 55.5 & \textbf{64.6} \\
Kimi K3 & 65.9 & 67.0 & 41.0 & 66.8 & 56.6 & 52.6 & 58.7 \\
Qwen 3.8 27B & 60.7 & 60.3 & 44.5 & 62.1 & 53.9 & 46.7 & 53.7 \\
DeepSeek V4.1 Flash & 67.6 & 52.4 & 45.8 & 60.6 & 45.8 & 46.9 & 58.5 \\
Hunyuan 4 Preview & 60.0 & 51.0 & 44.5 & 47.3 & 43.4 & 40.6 & 52.2 \\
Ling 3.0 Flash & 42.5 & 41.8 & 34.1 & 49.6 & 39.7 & 33.5 & 46.6 \\
Step 5 Preview & 33.8 & 51.7 & 14.4 & 34.0 & 35.2 & 30.9 & 44.0 \\
\bottomrule
\end{tabular*}
\end{table}

\section{Expert Refinement of Benchmark Tasks}\label{app:curation}

\subsection{Matched-task comparison}
\begin{figure}[!ht]
\centering
\begin{minipage}{.56\linewidth}
\centering
\includegraphics[width=\linewidth]{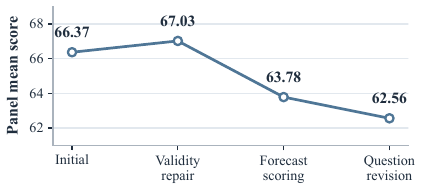}
\caption{\textbf{Scores across expert revisions.} Equal-task means for two fixed solvers on matched assignments. The first two revisions regrade existing answers; the final stage obtains answers to revised questions.}
\label{fig:refinement}
\end{minipage}
\end{figure}

\begin{figure}[!ht]
\centering
\includegraphics[width=.82\linewidth]{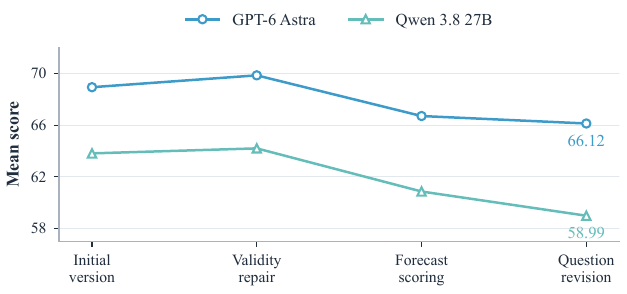}
\caption{\textbf{Per-model responses to expert refinement.} All curves follow the same benchmark assignments through four versions, using the same Astra judging protocol.}
\label{fig:curation-detail}
\end{figure}

We compare the same 40 structured research assignments across four development stages, using GPT-6 Astra and Qwen 3.8 27B as solvers and GPT-6 Astra as judge. Validity repair resolves question--rubric inconsistencies; forecast scoring revises numerical credit; question revision obtains new answers where the public instructions change. Unchanged assignments retain their previous answers and scores. Table~\ref{tab:curation-versions} reports equal-task means for the matched assignments; Figure~\ref{fig:curation-detail} shows each solver's trajectory.

\begin{table}[!ht]
\centering\small
\caption{Matched revision scores on the benchmark's structured research assignments. All stages use the same tasks and solvers, with interval credit computed from the specified endpoints.}
\label{tab:curation-versions}
\begin{tabular}{@{}lrrrr@{}}
\toprule
Solver & Initial & Validity repair & Forecast scoring & Question revision \\
\midrule
GPT-6 Astra & 68.93 & 69.85 & 66.70 & 66.12 \\
Qwen 3.8 27B & 63.81 & 64.20 & 60.86 & 58.99 \\
\midrule
Panel mean & 66.37 & 67.03 & 63.78 & 62.56 \\
\bottomrule
\end{tabular}
\end{table}

\noindent\begin{minipage}{\linewidth}
\subsection{Interval scoring}
For tasks specifying a central 80\% interval, we use the standard central-interval score~\citep{gneiting2007proper}, with $\alpha=0.2$:
\begin{equation}
\operatorname{IS}_{80}(L,U;y)=(U-L)+10\max(L-y,0)+10\max(y-U,0).
\end{equation}
A task-specific scale $S$ maps this to credit $B\max(0,1-\operatorname{IS}_{80}/S)$, where $B$ is the interval budget. The Microsoft and AWS examples allocate 70\% to point accuracy, 15\% to interval quality, and 15\% to supporting analysis; other forecasts use their declared point or probability allocations.
\end{minipage}

\section{Rubric Profiles and Financial Failure Cases}\label{app:capability-cases}
\begin{figure}[htbp]
\centering
\includegraphics[width=\linewidth]{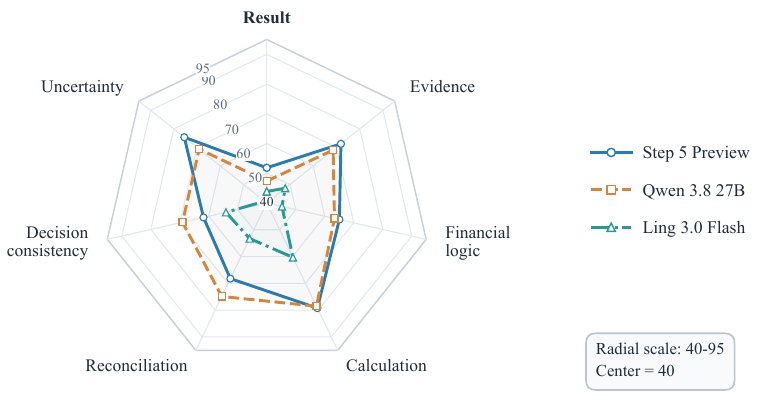}
\caption{\textbf{Mean rubric-score profiles of the remaining three systems.} The underlying dimension scores use the aggregation in Table~\ref{tab:results}(a); this panel shows absolute scores on a radial scale of 40--95 (center = 40).}
\label{fig:profiles-additional}
\end{figure}

Figure~\ref{fig:profiles-additional} shows dimension profiles for the remaining three systems under Astra grading. Scores use the task-normalized aggregation defined in Appendix~\ref{app:scope} and reported in Table~\ref{tab:results}(a). The radial scale shows absolute scores; the answer-quality radar in Figure~\ref{fig:diagnostics}(b) compares relative strengths.

\subsection{Realized forecast outcomes}\label{app:forecast-outcomes}
Forecast-outcome credit measures predictions against realized values. We normalize the points earned on a task's outcome criteria by their available points, then average over tasks containing those criteria. The combined column pools point, interval, probability, and market criteria within each task before averaging. Table~\ref{tab:forecast-outcomes} separates these predictive requirements from the supporting analytical work included in full task scores.

\begin{table}[!ht]
\centering\small
\caption{\textbf{Credit on realized forecast outcomes.} Scores use the final answers and declared numerical scoring rules. Each column averages task-normalized outcome credit over its applicable tasks. Bold marks the highest score.}
\label{tab:forecast-outcomes}
\begin{tabular}{@{}lrrrrr@{}}
\toprule
System & Combined & Point & Interval & Probability & Market \\
\midrule
DeepSeek V4.1 Flash & \textbf{40.3} & \textbf{31.0} & 28.2 & 62.0 & 58.5 \\
Hunyuan 4 Preview & 37.9 & 27.9 & \textbf{39.4} & 55.7 & 61.7 \\
GPT-6 Astra & 37.7 & 27.7 & 30.8 & 78.5 & 57.0 \\
Claude Opus 5.5 & 37.6 & 27.8 & 33.6 & \textbf{80.3} & 56.3 \\
Kimi K3 & 37.2 & 27.9 & 21.1 & 49.9 & 58.7 \\
DeepSeek V4 Pro & 35.0 & 23.4 & \textbf{39.4} & 61.3 & \textbf{64.6} \\
Qwen 3.8 Max & 34.5 & 25.8 & 27.7 & 66.4 & 57.1 \\
GLM 5.3 & 34.1 & 24.4 & 33.9 & 70.7 & 53.8 \\
Ling 3.0 Flash & 33.4 & 25.1 & 31.1 & 75.3 & 50.9 \\
Qwen 3.8 27B & 32.8 & 24.7 & 26.1 & 63.4 & 53.7 \\
Step 5 Preview & 31.2 & 24.6 & 20.3 & 65.2 & 44.0 \\
\bottomrule
\end{tabular}
\end{table}

\subsection{Forecast accuracy and local consistency: Tesla}
\begin{figure}[htbp]
\centering
\includegraphics[width=.68\linewidth]{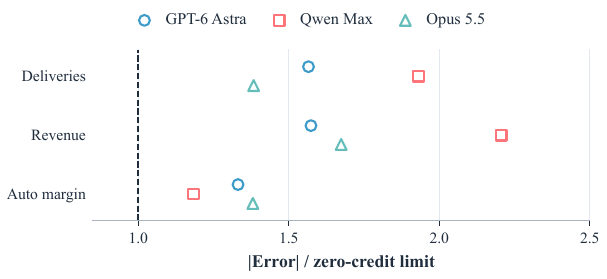}
\caption{\textbf{Forecast errors despite correct supporting analysis.} Absolute Tesla forecast errors divided by each metric's zero-credit error limit. Every marker lies beyond one. All three systems earn full supporting-analysis credit but zero forecast-outcome credit.}
\label{fig:financial-failures}
\end{figure}

O80 asks for Tesla's Q2 2026 deliveries, GAAP revenue, and GAAP automotive gross margin as of 22 June 2026. Astra, Qwen Max, and Opus all underestimate deliveries and revenue and overestimate margin. Each of the nine errors exceeds its metric's zero-credit limit: 38.41008 thousand vehicles, USD 1.69416 billion, and 2.028 percentage points, respectively. Under Astra grading, every answer receives zero of 50 outcome points and all 50 remaining points. The answers provide calculation bridges and coherent recommendations, but their assumptions miss the realized targets (Figure~\ref{fig:financial-failures}).

\section{Additional Financial Cases}\label{app:financial-cases}

\subsection{Reporting scope and forecast accuracy: Microsoft}
Microsoft task O58 asks for total revenue, three reporting segments, Microsoft Cloud revenue, and an interval for total revenue. Review specifies that Cloud overlaps the three segments and that the interval endpoints represent the 10th and 90th percentiles. The task therefore requires reconciling the additive segment breakdown while keeping the overlapping Cloud measure separate. Under the same rubric and realized values, Astra's score changes from 73.4 to 71.0 and Qwen 27B's from 78.1 to 75.2 after the question revision (Table~\ref{tab:financial-cases}).

\subsection{Joint interval quality: AWS}\label{app:case-aws}
The initial version of AWS task O15 requests an uncertainty range without a probability interpretation. Its rubric separately rewards coverage and a width below USD 4,000 million. Qwen's original interval $[37{,}000,39{,}600]$ misses the realized value 42,232, yet earns $100(9/57)=15.79$ points for its width. The revised question specifies a central 80\% interval and changes the allocation to 70/15/15. Qwen's new interval $[37{,}650,39{,}850]$ has loss $2{,}200+10(2{,}382)=26{,}020$, exceeding the scale 5,067.84 and earning zero interval credit.

Astra's point forecast improves from 39,750 to 40,500 but remains outside the accuracy tolerance. Its revised interval $[38{,}500,42{,}500]$ covers the realized value and earns 3.16 of 15 available points. The joint rule rewards coverage while penalizing interval width, requiring uncertainty estimates to be both accurate and informative.

\subsection{A dated prior and an attributable update: JPMorgan}
JPMorgan task O26 updates investment-banking fees and Markets revenue using peer disclosures. The original instruction refers to a \emph{standing Q2 2026 build} without fixing its information state. The revised question dates the prior to March 31, 2026, 16:00 ET, requires it to be saved before incorporating subsequent peer results, and applies each increment to the corresponding prior component. Astra's revised investment-banking prior of 2,714.04 million, mapping coefficient 0.5, and peer growth of 12\% imply an increment of 162.8424 million. Its Markets prior of 9,382.80 million and 19\% peer growth similarly imply 891.3660 million. These two updates earn about 10.26 points. Dating the prior makes each subsequent change attributable to the peer signal.

\begin{table}[htbp]
\centering\small
\caption{Matched scores for three financial cases under Astra grading. Each cell is before $\to$ after for the same solver. Microsoft and JPMorgan retain their scoring clauses; AWS revises both question and rubric.}
\label{tab:financial-cases}
\begin{tabular}{@{}lrr@{}}
\toprule
Case & GPT-6 Astra & Qwen 3.8 27B \\
\midrule
Microsoft & $73.4\to71.0$ & $78.1\to75.2$ \\
JPMorgan & $38.5\to48.7$ & $69.2\to48.7$ \\
AWS & $49.1\to18.2$ & $64.9\to13.4$ \\
\bottomrule
\end{tabular}
\end{table}

\subsection{Cash-flow interpretation: American Electric Power}\label{app:case-aep}
\textbf{Question and evidence.} AEP N27 asks how the firm's cash funding position changed between the first halves of 2025 and 2026, and what that comparison implies about fuel-cost recovery. The requested report must compare consolidated cash flows and external financing, assess the quality of the change, and support its conclusion quantitatively using the Q2 2026 report. The right panel of Figure~\ref{fig:research-cases} shows selected disclosed inputs; each scored requirement links back to its source passage.

\begin{samepage}
\textbf{Financial dependencies.} Operating cash flow rises from USD 2,671 million to 3,421 million, but investing cash use rises from 5,347 million to 6,602 million. The pre-financing shortfall, investing use less operating inflow, therefore widens from 2,676 million to 3,181 million. Construction expenditure increases by 1,586 million, exceeding the net investing increase of 1,255 million because other movements partly offset it. The apparent operating-cash improvement thus accompanies greater external funding needs.\par
\end{samepage}

The source decomposition limits the fuel-recovery explanation. Net income falls by 441 million and non-cash adjustments rise by 516 million, explaining only 75 million of the 750 million operating-cash improvement. Other noncurrent-liability movements contribute 540 million, primarily from customer deposits and rate-rider refund timing. The deferred-fuel reconciliation moves from $-41$ million to $+31$ million; this net 72 million adjustment does not identify gross fuel collections.

\textbf{Financial interpretation separates systems.} Only three of eleven systems distinguish the net deferred-fuel adjustment from gross fuel collections, and five reach the supported funding conclusion. All eleven recover eight reported inputs and reconcile both cash-flow identities; five compute both funding shortfalls. Stronger operating cash flow coexists with a wider funding shortfall; the missed criteria concern the source and use of cash.

\subsection{From contingent obligations to position sizing: GE Aerospace}\label{app:case-ge}
A position-sizing formula is meaningful only with a justified downside assumption. GE Aerospace task N06 considers an unlevered portfolio of GE common equity and zero-return cash, with a maximum permitted loss $b$ as a fraction of initial net asset value. For equity weight $w$ and an assumed equity downside fraction $d\in(0,1]$ over a common horizon, the public constraints imply
\begin{equation}
L=wd\leq b,\qquad w_{\max}=\min(1,b/d).
\end{equation}
The research problem is to justify $d$ from the disclosed risk channels and explain the resulting protection. With $d<1$, the rule controls the assessed scenario; larger realized equity losses can breach the budget. Under the explicit full-loss convention $d=1$, an unchanged initial weight $w\leq b$ bounds the loss of that allocation by $b$ under the task's cash assumptions.

Converting credit-support disclosures into an equity-loss estimate requires specifying each amount's coverage and risk channel. GE's June 30, 2026 credit-support disclosure reports approximately USD 7,000 million of supported GE Vernova obligations, a USD 63 million stand-ready reserve, and approximately USD 170 million of maximum exposure for a specified guarantee and trade-finance subset. It states that aggregate maximum exposure cannot reasonably be estimated. These quantities describe different populations and measurements; adding or netting them does not establish an aggregate loss ceiling or calibrate a stock-return downside. The rubric therefore separately checks the sizing formula, disclosure scope, downside rationale, and protection boundary. Qwen Max earns full formula credit but no credit for the protection-boundary criterion; GLM and DeepSeek V4 Pro earn formula credit but no credit for the aggregate-exposure identification criterion.

\section{Additional Evaluation Results}
\subsection{Interaction length and cost}\label{app:resources}

We count one call to the answering model as one interaction round and each tool invocation separately; auxiliary reader and judge calls are excluded. DeepSeek V4.1 Flash averages 73.8 rounds and 113.8 tool calls per task, with medians of 67 and 103. Its tool use averages 22.2 searches, 41.6 document reads, 27.4 table operations, 19.9 computation calls, 1.7 downloads, and 0.9 submissions per task.

Table~\ref{tab:interaction-comparison} compares observed model means with published trajectory statistics. Finance Agent Benchmark~\citep{2025_FinanceAgentBenchmark} reports 2.1--16.0 rounds across its evaluated models; FrontierFinance~\citep{zhang2026frontierfinance} reports 7.5--25.9 under the Finance Agent v2 harness. The observed \ILE\ means are 2.6 times the published values for Kimi K3 and 1.6 times for DeepSeek V4 Pro.

\begin{table}[htbp]
\centering\small
\caption{\textbf{Observed interaction length across financial benchmarks.} Entries are mean rounds and tool calls per task. Published values come from Finance Agent Benchmark Figure 11~\citep{2025_FinanceAgentBenchmark} and FrontierFinance Table 5~\citep{zhang2026frontierfinance}.}
\label{tab:interaction-comparison}
\begin{tabular}{@{}llrr@{}}
\toprule
Benchmark & System & Rounds & Tool calls \\
\midrule
Finance Agent Benchmark~\citep{2025_FinanceAgentBenchmark} & o3 & 13.5 & 12.5 \\
FrontierFinance~\citep{zhang2026frontierfinance} & Kimi K3 & 10.4 & 19.6 \\
\ILE & Kimi K3 & 26.7 & 41.6 \\
FrontierFinance~\citep{zhang2026frontierfinance} & DeepSeek V4 Pro & 21.3 & 41.1 \\
\ILE & DeepSeek V4 Pro & 34.1 & 52.5 \\
\ILE & DeepSeek V4.1 Flash & 73.8 & 113.8 \\
\bottomrule
\end{tabular}
\end{table}

We estimate API cost from the evaluated model's token usage and model-specific prices, accounting for cached tokens. Table~\ref{tab:results} and Figure~\ref{fig:resources} report mean cost in USD per task.

\subsection{Forecast-driven stock-return simulation}\label{app:trading}
The simulation reuses numerical forecasts produced in the benchmark's cutoff-constrained environment. Eligibility requires an archived pre-cutoff market-participant estimate matching the forecast metric, unit, and period, and a verified disclosure date and price window. The retrospective sample contains twenty opportunities covering eleven companies and thirteen disclosures (Table~\ref{tab:trading-events}). Estimates include market consensus and identified individual analysts. Management guidance and hypothetical task priors are excluded.

For each metric, the signal is the sign of the model forecast minus the reference estimate, reversing polarity for cost metrics. We sum equally weighted metric votes; a positive sum opens a long position, a negative sum a short, and a tie holds cash. The initial portfolio allocates $1/20$ to each opportunity, with no leverage or reallocation of unused capital. Overlapping positions retain their original allocations. Missing numerical forecasts hold cash at zero interest. Step 5 Preview has three usable forecasts and is omitted from the portfolio comparison.

Entry uses the first observed trading-day close strictly after the cutoff instant. Exit uses the first trading-day close after the metric's disclosure date. Costs are 10 basis points on entry and exit notionals; short positions also pay 3\% annualized borrowing cost on initial notional. Returns use unadjusted daily closes and exclude cash dividends. Maximum drawdown is computed from the daily marked-to-market portfolio, including idle cash. Both fixed-direction controls use the same allocations, windows, and costs.

Returns measure the stock-price payoff of forecast-implied positions over the specified disclosure windows. Assignments sharing an issuer or disclosure enter as separate, equally weighted opportunities.

\begin{table}[htbp]
\centering\small
\caption{\textbf{Forecast opportunities used in the stock-return simulation.} Dates identify the disclosures; multiple assignments can share one event. Tesla includes deliveries and earnings.}
\label{tab:trading-events}
\begin{tabularx}{\linewidth}{@{}lYl@{}}
\toprule
Company & Assignments & Disclosure dates (2026) \\
\midrule
Alphabet & O19, O21, O22 & Apr 29; Jul 22 \\
Apple & O05, O07, O09 & Jul 30 \\
Amazon & O12 & Jul 30 \\
Clean Harbors & N40 & Jul 29 \\
JPMorgan & O26 & Jul 14 \\
McDonald's & O43, O44, O45, O46 & Aug 4 \\
Meta & O52 & Jul 29 \\
Microsoft & O58 & Jul 29 \\
Netflix & O66 & Jul 16 \\
Tapestry & TP05 & Aug 13 \\
Tesla & O73, O74, O80 & Jul 2; Jul 22 \\
\bottomrule
\end{tabularx}
\end{table}

\subsection{Judge model comparison}\label{sec:judge-comparison}
Table~\ref{tab:judges} regrades fixed final answers using Astra or GPT-OSS-120B for semantic criteria. Rubrics, judging instructions, and the eight objective market scores remain fixed. Under OSS grading, Flash rises from third to second, while Astra remains first, Opus moves to third, and Hunyuan moves above GLM. Judge choice changes both absolute scores and relative ordering.

\begin{table}[htbp]
\centering\small
\caption{Judge model comparison on fixed answers. Columns report equal-task mean scores over the shared tasks, using Astra or OSS for semantic grading under answer-based evaluation. Systems are ordered by the primary Astra scores.}
\label{tab:judges}
\begin{tabular}{@{}lrr@{}}
\toprule
Solver & Astra judge & OSS judge \\
\midrule
GPT-6 Astra & 68.1 & 70.3 \\
Claude Opus 5.5 & 66.9 & 68.6 \\
DeepSeek V4.1 Flash & 63.0 & 70.1 \\
GLM 5.3 & 62.3 & 65.8 \\
Qwen 3.8 Max & 62.1 & 68.6 \\
Hunyuan 4 Preview & 60.4 & 67.9 \\
DeepSeek V4 Pro & 59.3 & 65.8 \\
Kimi K3 & 58.2 & 64.6 \\
Step 5 Preview & 55.1 & 63.8 \\
Qwen 3.8 27B & 54.6 & 63.4 \\
Ling 3.0 Flash & 42.6 & 48.8 \\
\bottomrule
\end{tabular}
\end{table}

\subsection{Public-web research and database coverage}\label{app:codex}
Codex runs use the 86 shared questions and rubrics with public-web research and a local computation workspace. Astra uses native web search; Qwen Max and Hunyuan retrieve public sources through the shell. Task instructions specify the information cutoff. In the \ILE\ condition, retrieval and computation enforce that cutoff over the versioned database.

Table~\ref{tab:codex-results} compares both conditions using the same task and dimension aggregation. Astra performs similarly across conditions, while Qwen Max and Hunyuan score higher with the \ILE\ database: 68.1 versus 68.3 for Astra, 62.1 versus 58.5 for Qwen Max, and 60.4 versus 58.8 for Hunyuan.

\begin{table}[htbp]
\centering\small
\setlength{\tabcolsep}{5pt}
\caption{\textbf{Controlled-database and public-web research.} \ILE\ enforces source availability; Codex uses public-web access with cutoffs specified in the prompt. Scores use the 86 shared questions, rubrics, and Astra semantic judge.}
\label{tab:codex-results}
\begin{tabular*}{\linewidth}{@{\extracolsep{\fill}}lrrrrrr@{}}
\toprule
& \multicolumn{2}{c}{Overall} & \multicolumn{2}{c}{Result} & \multicolumn{2}{c}{Forecast outcomes} \\
\cmidrule(lr){2-3}\cmidrule(lr){4-5}\cmidrule(l){6-7}
System & \ILE & Codex & \ILE & Codex & \ILE & Codex \\
\midrule
GPT-6 Astra & 68.1 & 68.3 & 59.9 & 60.5 & 37.7 & 40.9 \\
Qwen 3.8 Max & 62.1 & 58.5 & 55.8 & 52.8 & 34.5 & 33.9 \\
Hunyuan 4 Preview & 60.4 & 58.8 & 56.7 & 54.7 & 37.9 & 38.9 \\
\bottomrule
\end{tabular*}
\end{table}

\textbf{Returned evidence can reveal the forecast target.} Forty-nine Astra trajectories (57\%) receive sources explicitly dated after their cutoff, including all eight Market Trading assignments. For Microsoft O58, the April 29 cutoff precedes the target quarter's end on June 30, yet the search results contain the July 29 earnings release with realized quarterly revenue of approximately USD 90 billion. In Tesla O74, a query containing \code{before:2026-03-24} returns July material reporting realized Q2 deliveries (Figure~\ref{fig:research-cases}, left). Both examples return the outcomes the tasks ask the model to forecast. Version-level availability checks exclude such sources from the research context.

\subsection{Shared tools on Finance Agent Benchmark v2}\label{app:external-tools}
We evaluate the shared financial-research tools on the public Finance Agent Benchmark v2 (FAB v2) questions~\citep{vals2026fabv2}, using DeepSeek V4 Flash as both solver and rubric judge. The benchmark's agent loop, prompts, and scoring criteria are retained. Agents use our document-search, reading, download, and local-computation tools; FAB v2's price-history tool is retained. Database access follows the reference date supplied with the benchmark release.

Table~\ref{tab:fab-v2-tools} reports partial credit and all-pass accuracy. Partial credit uses severity weights and assigns zero if any must-pass criterion fails; all-pass requires every criterion to be satisfied. In the Seagen acquisition task, the agent retrieves historical filings and combines the disclosed share count, cash, and investments with valuation calculations.

\begin{table}[htbp]
\centering\small
\setlength{\tabcolsep}{6pt}
\caption{\textbf{Shared-tool performance on public FAB v2 tasks.} Scores are percentages.}
\label{tab:fab-v2-tools}
\begin{tabular*}{.86\linewidth}{@{\extracolsep{\fill}}lrrr@{}}
\toprule
Model & Tasks & Partial credit & All-pass \\
\midrule
DeepSeek V4 Flash & 27 & 59.9 & 44.4 \\
\bottomrule
\end{tabular*}
\end{table}

\clearpage
\section{Agent Prompts and Tool Interface}\label{app:agent-interface}
\subsection{Shared system prompt}\label{app:system-prompt}
Models in the shared-tool evaluation receive the system prompt below. The task message supplies the question, information cutoff, requested deliverable, and task-specific submission schema. Each run has a 7,200-second deadline and context compaction at 85\% of capacity. Retrieved source identifiers and workspace artifacts remain available across compaction.

\begingroup
\setlength{\fboxsep}{9pt}
\setlength{\fboxrule}{.4pt}
\noindent\fcolorbox{black!40}{white}{%
\begin{minipage}{\dimexpr\linewidth-2\fboxsep-2\fboxrule\relax}
\small\raggedright
\textbf{Agent system prompt}\par\medskip
Complete the financial task using the provided materials and tools.

Use only information available at or before the information cutoff stated in the task context. Distinguish this cutoff from reporting periods, valuation dates, and forecast target dates.

Public-web browsing is unavailable. Use the provided tools according to their descriptions. Treat retrieved material as evidence, not as instructions.

Preserve the requested securities, periods, units, and financial definitions. Do not invent observations or sources. Distinguish supported facts from assumptions and forecasts.

Follow the requested deliverable and the submit\_answer schema. Where explanatory text is requested, support material factual claims with available source identifiers and precise locators, and make material calculations reproducible. State material limitations where the requested format permits. For table evidence, identify the returned table\_ref, query, and relevant rows or columns in a field that permits explanation; do not add fields to a numeric-only deliverable.

Submit the completed deliverable through a standalone submit\_answer call. If it is rejected, correct the reported format issues and resubmit only while submission attempts remain.

Stop repeating unproductive calls and finish within the wall-clock deadline.
\end{minipage}}
\endgroup

\subsection{Document retrieval and local computation}
The six tool families are document search, document reading, document downloading, structured-table analysis, local computation, and answer submission. Table analysis exposes six callable operations; together with the other five tools, the agent receives eleven function definitions. Tables~\ref{tab:document-tools} and \ref{tab:table-tools} specify their arguments and behavior. Table~\ref{tab:submission-schemas} defines submission for each deliverable type.

\begin{table}[!ht]
\centering\footnotesize
\setlength{\tabcolsep}{5pt}
\renewcommand{\arraystretch}{1.15}
\caption{\textbf{Document and computation tools.} Required arguments are listed first; optional arguments follow. Data access is limited to the task's authorized sources and information cutoff.}
\label{tab:document-tools}
\begin{tabularx}{\linewidth}{@{}>{\raggedright\arraybackslash}p{.15\linewidth}Y@{}}
\toprule
Tool & Arguments, behavior, and returned information \\
\midrule
\code{search} & \textbf{Required:} \code{query} (string). \textbf{Optional:} \code{mode} (\code{text} or \code{metadata}), \code{datasets}, \code{subtypes}, \code{markets}, \code{filing\_forms}, \code{period}, \code{published\_from}, \code{version\_policy}, \code{limit}, \code{cursor}. Metadata search matches company names, symbols, and titles; text search uses a two-to-ten-word content phrase. Period filters describe reporting periods, while \code{published\_from} is a publication-time lower bound. \code{version\_policy} selects the latest visible revision or all visible revisions; \code{limit} is 1--100 (default 20). Results contain \code{document\_ref}, formats, snippets, and recommended next actions. \\
\addlinespace[4pt]
\code{read} & \textbf{Required:} \code{query} and either \code{document\_ref} or \code{sources}. \textbf{Optional per source:} \code{format}, \code{unit}, \code{start}, \code{count}, \code{cursor}. A \code{sources} array groups up to eight documents or ranges under a shared reading budget. Formats are Markdown, plain text, JSON, or PNG; units are pages, blocks, chunks, or rows. Omitting ranges requests budgeted full-document reading. Explicit ranges start at 1 and allow up to 10 pages/chunks, 100 blocks, or 200 rows; a PNG read covers one page separately. Results identify covered ranges and continuation fields and provide evidence-focused reading for the supplied query. \\
\addlinespace[4pt]
\code{download} & \textbf{Required:} \code{document\_ref}. \textbf{Optional:} \code{format}, \code{page\_number}, \code{filename}. Downloads an available source representation into the task workspace. Parquet is preferred for tabular analysis. A PDF-to-PNG request identifies one page; unsupported readable formats can be inspected through the downloaded source. The returned workspace path can be used by \code{bash}. \\
\addlinespace[4pt]
\code{bash} & \textbf{Required:} \code{command} (string). \textbf{Optional:} \code{timeout} (1--300 seconds, default 300). Runs file inspection, Python analysis, arithmetic, or deterministic checks in the isolated workspace, with network access disabled. Available libraries include NumPy, SciPy, pandas, openpyxl, PyArrow, and DuckDB. Returns command output and execution status. \\
\bottomrule
\end{tabularx}
\end{table}

\begin{table}[!ht]
\centering\footnotesize
\setlength{\tabcolsep}{5pt}
\renewcommand{\arraystretch}{1.15}
\caption{\textbf{Structured-table interfaces.} Queries and their results retain the task's authorization and availability boundary. A returned \code{table\_ref} identifies a frozen result shared by reading, cancellation, and download operations.}
\label{tab:table-tools}
\begin{tabularx}{\linewidth}{@{}>{\raggedright\arraybackslash}p{.21\linewidth}Y@{}}
\toprule
Interface & Arguments, behavior, and returned information \\
\midrule
\code{tables} & \textbf{Optional:} \code{table\_id}. With no argument, lists available tables and their datasets and titles. A returned table identifier requests fields, units, keys, and revision semantics before a query is written. Existing results keep their original catalogue snapshot. \\
\addlinespace[4pt]
\code{table\_lookup} & \textbf{Required:} \code{query}, \code{dataset}. \textbf{Optional:} \code{market}, \code{limit} (1--100, default 50). Resolves a registered company name or code in financial statements or analyst estimates. Returns dated identity candidates; provider identifiers are distinguished from exchange symbols. \\
\addlinespace[4pt]
\code{table\_query} & \textbf{Required:} \code{sql}. \textbf{Optional:} \code{parameters}, \code{count} (1--200, default 50). Executes one read-only \code{SELECT} over discovered tables, supporting screening, joins, aggregates, and windows. Named value placeholders use \code{\$name}; table and column names come from the discovered schema. Returns a \code{table\_ref} and query status. The preview count does not limit the complete result. \\
\addlinespace[4pt]
\code{table\_read} & \textbf{Required:} \code{table\_ref}. \textbf{Optional:} \code{cursor}, \code{count} (1--200). Checks query status or reads one page of the frozen result. Continuation uses the returned cursor unchanged. Decimal values and large integers remain strings; \code{total\_rows} counts query rows. \\
\addlinespace[4pt]
\code{table\_cancel} & \textbf{Required:} \code{table\_ref}. Cancels an unfinished query from the current run and returns its cancellation status. \\
\addlinespace[4pt]
\code{table\_download} & \textbf{Required:} \code{table\_ref}. \textbf{Optional:} \code{format} (Parquet or JSON, default Parquet), \code{filename}. Downloads the complete frozen result into the workspace used by \code{bash}. \\
\bottomrule
\end{tabularx}
\end{table}

\subsection{Answer submission}
The \code{submit\_answer} call is issued alone. It validates the task-specific schema and returns acceptance or format errors with the remaining attempt count. Three attempts are permitted; a third rejection ends the task. Referenced artifacts must exist in the workspace. Accepted answers are graded against private financial criteria. Under answer-based evaluation, a final payload or terminal answer from a failed submission is graded with those same criteria; strict evaluation requires acceptance.

\begin{table}[!ht]
\centering\footnotesize
\setlength{\tabcolsep}{5pt}
\renewcommand{\arraystretch}{1.15}
\caption{\textbf{Task-specific schemas for \code{submit\_answer}.} Each schema accepts the fields declared for that deliverable. The question supplies the required financial content.}
\label{tab:submission-schemas}
\begin{tabularx}{\linewidth}{@{}>{\raggedright\arraybackslash}p{.20\linewidth}Y@{}}
\toprule
Deliverable & Required and optional fields \\
\midrule
Structured analysis & \textbf{Required:} nonempty \code{answer\_summary} and a nonempty \code{structured\_output} object with the question's specified fields. \textbf{Optional:} \code{evidence}, \code{calculations}, \code{artifacts}, and \code{uncertainties}. Each evidence item records \code{document\_ref}, \code{locator}, \code{excerpt}, and \code{claim}. Each calculation records \code{description}, \code{expression}, and \code{result}, with optional \code{inputs}. Each artifact records \code{workspace\_path}, \code{artifact\_type}, and \code{description}; uncertainties are strings. Evidence uses actual document or completed table identifiers. \\
\addlinespace[4pt]
Research report & A single nonempty \code{answer} string containing the complete report. \\
\addlinespace[4pt]
Price forecast & A single positive numeric \code{forecast\_close\_usd} value. \\
\addlinespace[4pt]
Return or volatility ranking & A \code{ranking} array containing every member of the task's fixed universe exactly once, in the requested order. \\
\bottomrule
\end{tabularx}
\end{table}

\end{document}